\documentclass{article}

\usepackage{microtype}
\usepackage{graphicx}
\usepackage{subcaption}
\usepackage{booktabs} % for professional tables

\usepackage{hyperref}

\usepackage[accepted]{icml2026}

\usepackage{amsmath}
\usepackage{amssymb}
\usepackage{mathtools}
\usepackage{amsthm}

\usepackage[capitalize,noabbrev]{cleveref}

\usepackage{wrapfig}
\usepackage{multirow}

\usepackage{xcolor}

\theoremstyle{plain}

\theoremstyle{definition}

\theoremstyle{remark}

\icmltitlerunning{Stability-Aware Feature Design for Robust Watermark Detection}

\begin{document}

\twocolumn[
\icmltitle{Stability-Aware Feature Design for Robust Watermark Detection in Machine-Generated Text}

% It is OKAY to include author information, even for blind submissions: the
% style file will automatically remove it for you unless you've provided
% the [accepted] option to the icml2026 package.

% List of affiliations
\icmlsetsymbol{equal}{*}

\begin{icmlauthorlist}
\icmlauthor{Sina Mansouri}{equal,gmu}
\icmlauthor{Mohit Marvania}{equal,gmu}
\icmlauthor{Abolfazl Safikhani}{gmu-stat}
\end{icmlauthorlist}

\icmlaffiliation{gmu}{Department of Computer Science, George Mason University, Fairfax, VA, USA}
\icmlaffiliation{gmu-stat}{Department of Statistics, George Mason University, Fairfax, VA, USA}

\icmlcorrespondingauthor{Abolfazl Safikhani}{asafikha@gmu.edu}

% You may provide any keywords that you find helpful for describing your paper
\icmlkeywords{Watermarking, Large Language Models, Machine-Generated Text Detection, Paraphrasing Robustness}

\vskip 0.3in
]

% this must go after the closing bracket ] following \twocolumn[ ...
\printAffiliationsAndNotice{\icmlEqualContribution}

\begin{abstract}
The widespread adoption of large language models (LLMs) has intensified the demand for principled methods to distinguish human from machine-generated text. Watermarking provides a promising avenue, yet existing detectors exhibit sharp performance deterioration under multiple paraphrasing and when applied to shorter texts. We introduce \textit{Pattern Stability Score (PSS)}, a novel detection framework that leverages local statistical features and stability dynamics across paraphrased variants. Specifically, the proposed method combines global and local z-score features with higher-order statistics of run-length patterns, enriched by autocorrelation signals and stability scores computed over paraphrase depth. Numerical evaluations are performed on three benchmark datasets (PG-19, CNN/DailyMail, and WikiText) using multiple LLMs (Llama-3-8B, Qwen2-7B) and paraphrasers (Mistral-7B, Qwen2-7B, Gemma-7B), systematically stress-testing robustness under up to eight rounds of paraphrasing. Compared to prior z-score thresholding baselines and some state-of-the-art deep learning methods, our approach improves detection AUC (area under the receiver operating characteristic curve) by over 10-15 percentage points across different token lengths. Additionally, extensive cross-domain experiments demonstrate that a single universal classifier generalizes across different LLMs, paraphrasers, and text domains without retraining, maintaining above 87.8\% AUC even when all components differ from training.
%77\% accuracy even when all components differ from training. 
\end{abstract}

\section{Introduction}

% Further, at the critical 1\% false positive rate threshold, PSS achieves 84\% true positive rate compared to only 42\% for global z-score methods, a 2x improvement essential for minimizing false accusations in practical deployments. Additionally, it achieves strong precision-recall balance and AUC greater than 0.95 at full length, demonstrating resilience where prior detectors collapse.

% Finally, sensitivity analysis is conducted on window size, stride, and token length to validate design choices. Overall, these empirical results establish PSS as a practical and extensible framework for watermark detection, highlighting stability-based features as a promising direction for safeguarding LLM outputs against potential adversarial paraphrasing.

Large language models (LLMs) are now deployed at scale across both consumer and enterprise applications. As they increasingly integrate into writing workflows, the need to identify machine-generated content has shifted from a primarily academic inquiry to a practical requirement across domains such as education \cite{susnjak2022chatgpt,cotton2023chatting}, journalism \cite{chen2024can,zhou2023synthetic}, and science policy \cite{nas2024protecting,gao2023comparing}, among others. A long history of \emph{post-hoc} detection has been explored. For example, GLTR \cite{linguistic_features} leverages rank histograms from a reference LM to highlight text that disproportionately employs high-probability tokens. This approach is efficient but its reliance on rank features renders it vulnerable to paraphrasing and domain variation. Similarly, Grover \cite{zellers2019grover} jointly trains a generator--discriminator pair, using an in-domain classifier for detection. However, the performance declines when either the generator or domain changes, with paraphrasing further diminishing robustness. More recent work, such as Binoculars \cite{hans2024binoculars}, compares likelihoods under two open LMs and applies a likelihood-ratio style criterion for zero-shot detection. This improves cross-domain generalization but still exhibits sensitivity to paraphrasing and short inputs. Other methods, including curvature and rank-based tests such as DetectGPT and its variants \cite{detectgpt}, similarly rely on probability access from one or more LMs and remain susceptible to paraphrase smoothing.

In contrast, our focus is on \emph{watermarking}, which offers several distinctive advantages: it enables a keyed hypothesis test with controllable false positive rates under the null, a level of statistical precision difficult to attain with learned post-hoc classifiers whose error rates shift across domains, requires only token identities on the detector side (removing dependence on proprietary probability distributions), and anchors attribution in the provider's secret key rather than in learned stylistic features. At a high level, watermarking operates as follows: the generator biases token selection toward a hidden ``greenlist'' so that downstream text exhibits detectable statistical structure, while remaining human-readable \cite{kirchenbauer2023watermark,qu2025provably,he2025theoretically,lau-etal-2024-waterfall}. While alternative watermarking families have been proposed, including distortion-free schemes via inverse-transform sampling \cite{kuditipudi2024robust}, exponential-sampling approaches \cite{aaronson2022watermarking}, and cryptographically undetectable watermarks \cite{christ2024undetectable}, our scope is the greenlist family since variants of it have been deployed at production scale \cite{dathathri2024scalable}.
The standard detector aggregates evidence into a \emph{global} z-score and compares it to a threshold. However, a determined adversary can paraphrase the text, diluting or locally rearranging this signal \cite{canreliablydetected,cheng2025adversarialparaphrasinguniversalattack,detectgpt,fastdetectgpt}. There exist several methods to modify the watermarking scheme to make it more robust with respect to certain adversarial attacks such as paraphrasing (see Section~\ref{sec:related} for a review on other types of watermarking schemes). These studies motivate our design choice: instead of changing the generator to resist paraphrasing, we change the detector to exploit signals that paraphrasing preserves only imperfectly, namely local structure and stability across rewrites.

This detector-centric approach offers five practical advantages over modifying watermarking schemes. First, it provides \emph{deployment compatibility}: simple greenlist schemes have already been adopted at production scale \cite{dathathri2024scalable}, and detector-only improvements require no changes to generation pipelines, model architectures, or serving infrastructure. Second, it enables \emph{retroactive applicability}: enhanced detectors can identify watermarks in previously generated content without regeneration, providing immediate value for the vast corpus of already-watermarked text and for legal, educational, and journalistic settings where historical attribution matters. Third, it achieves \emph{robustness through defense-in-depth}: treating the watermark as a fixed signal and extracting evidence through multiple statistical lenses (local, global, stability-based) avoids the new attack surfaces and scheme-specific patterns that adversaries can exploit in more complex schemes \cite{semamark2023,saemark2024,diaa2025optimizing}. Fourth, it \emph{preserves generation quality and efficiency} without the stronger biases or auxiliary models required by multi-bit and adaptive schemes \cite{majormark2024,feng2025bimark}. Finally, it approaches the \emph{theoretical optimality limits} for fixed watermarking schemes through better statistical analysis \cite{statframework2024}, consistent with the principle of extracting all available information before declaring the need for stronger watermarks.

Recent works on watermarking repeatedly highlight two open gaps: robustness to \emph{multi-step paraphrasing} and stability on \emph{short texts} \cite{canreliablydetected,cheng2025adversarialparaphrasinguniversalattack}. To address these gaps, we study a black-box adversary who can paraphrase any given text up to $K$ steps using a strong instruction-tuned LLM. The adversary does \emph{not} know the watermark key or parameters and will preserve the original semantics and approximate length. Let $x^{(0)}$ denote the original passage (possibly watermarked) and ${x^{(k)}}_{k=1}^{K}$ its paraphrases at depths $D1{\ldots}D{K}$. The detector receives a single text at test time (any $x^{(k)}$ for $k=0,1, \ldots, K$). Our objective is to maintain detection power under paraphrasing and across different text lengths while controlling false positives on human-written content.

\paragraph{Why global z-score fails under paraphrasing.}
Greenlist watermarking \cite{kirchenbauer2023watermark} induces a binary indicator sequence over tokens (green/non-green) and a corresponding z-score measuring deviation from the null. The binary indicator sequence is constructed as follows: for each token $s^{(t)}$ in the generated text, we assign a value of 1 if the token belongs to the green list $G^{(t)}$ and 0 if it belongs to the red list $R^{(t)}$. Specifically, at each position $t$, a hash function seeded by the previous token $s^{(t-1)}$ deterministically partitions the vocabulary into a green list of size $\gamma|V|$ and a red list of size $(1-\gamma)|V|$, where $\gamma$ is typically 0.25 or 0.5. During watermarked generation, tokens from the green list are softly promoted by adding a bias $\delta$ to their logits. At detection time, we reconstruct this binary sequence by checking whether each observed token $s^{(t)}$ falls in its corresponding green list $G^{(t)}$ (assigned 1) or red list $R^{(t)}$ (assigned 0), using the same hash function and seed. Global tests summarize all tokens into one statistic (e.g. z-score) that depends only on the total green-token count, not on where green tokens appear in the sequence. Paraphrasing replaces some green tokens with non-green tokens, and crucially, this replacement is \emph{spatially non-uniform}: some regions retain strong watermark signals while others are heavily edited. The global $z$-score averages over the full sequence, so concentrated evidence in surviving regions is diluted by the heavily-edited regions, reducing the aggregate statistic even when localized evidence remains strong. A local detector, in contrast, can recover this heterogeneous signal by identifying windows where watermark evidence concentrates, the core motivation for our approach. The failure is structural: a single aggregate discards \emph{where} evidence concentrates and \emph{how} it behaves under edits.

% Global tests summarize all tokens into one statistic (e.g. z-score). Paraphrasing can break up long runs of green tokens, relocate them, or introduce local pockets of off-green context. These operations reduce the global statistic even when local evidence remains strong. The failure is structural: a single aggregate discards \emph{where} evidence concentrates and \emph{how} it behaves under edits.

% \begin{figure}[t]
% \vskip 0.2in
% \begin{center}
% \centerline{\includegraphics[width=0.95\columnwidth]{figures/Pipeline_updated_final.png}}
% \caption{\textbf{End-to-end PSS pipeline.} Human and AI-watermarked texts undergo up to eight paraphrasing rounds, with binary sequences extracted at each iteration (D0--D8). Watermarked text maintains stable patterns across iterations while human text shows random variation. An XGBoost classifier uses these stability features alongside static features for final classification.}
% \label{fig:pipeline}
% \end{center}
% \vskip -0.2in
% \end{figure}
\begin{figure*}[t]
\vskip 0.2in
\begin{center}
\centerline{\includegraphics[width=0.95\textwidth]{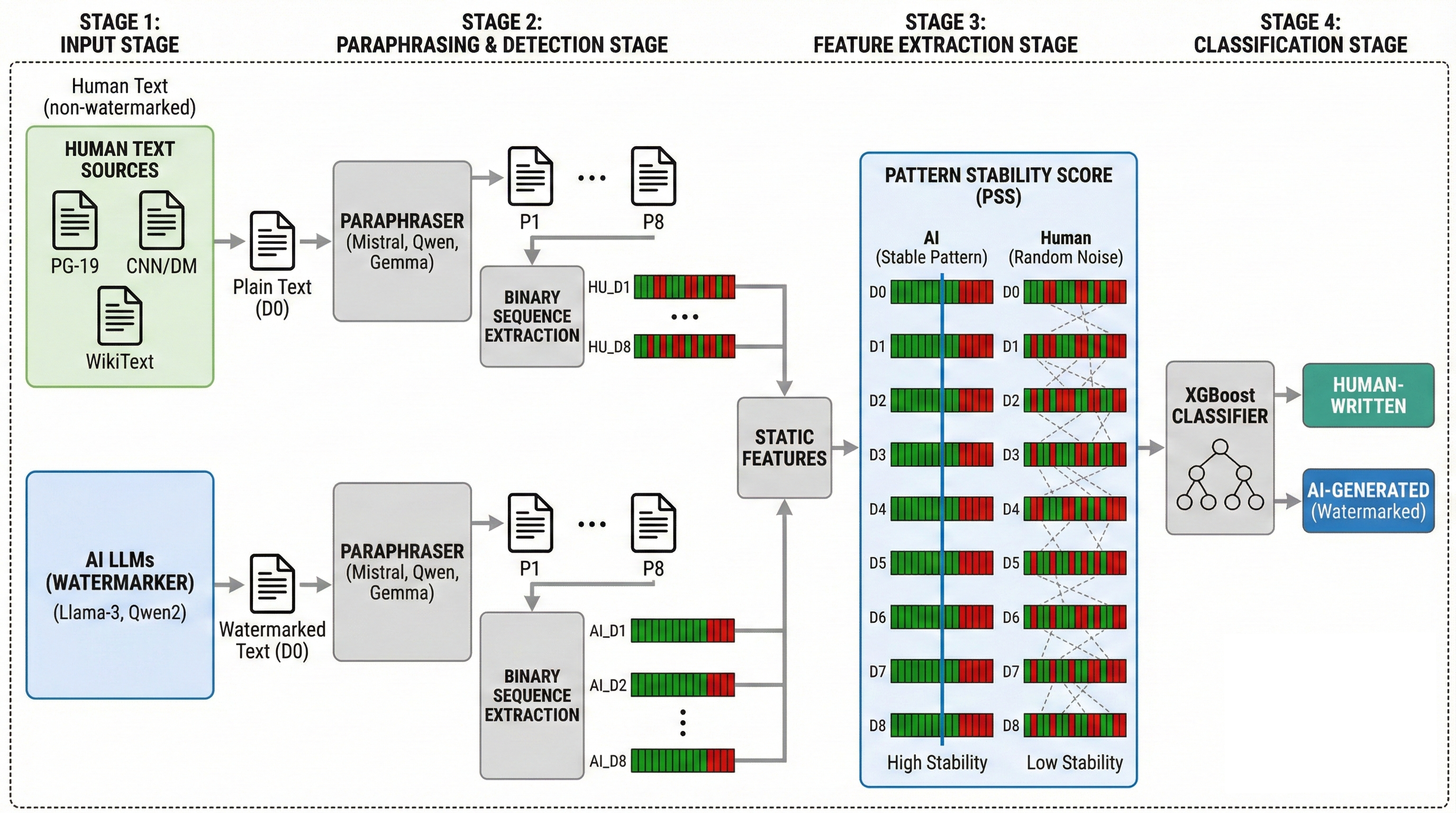}}
\caption{\textbf{End-to-end PSS pipeline.} Human and AI-watermarked texts undergo up to eight paraphrasing rounds, with binary sequences extracted at each iteration (D0--D8). Watermarked text maintains stable patterns across iterations while human text shows random variation. An XGBoost classifier uses these stability features alongside static features for final classification.}
\label{fig:pipeline}
\end{center}
\vskip -0.2in
\end{figure*}

\paragraph{Key idea: stability-aware local detection.}
The proposed method \textit{Pattern Stability Score (PSS)}, is a detection framework that (i) extracts \emph{local} watermark evidence via a rolling window and (ii) quantifies \emph{stability} of that evidence across paraphrase depth. Specifically, we slide a window along the 0-1 sequence and compute several local statistics within each window. Note that when a window splits a consecutive sequence/run of ones, we expand it minimally so runs are not fragmented while shrinking the tail window to cover all remaining tokens. For each window we compute a 20-dimensional feature set: six summary statistics of the z-score sequence across windows (mean, variance, min, max, skew, and kurtosis), lag-1 and lag-2 autocorrelations of z-scores, six summary statistics of longest-run length in the binary sequence as well as the same six summary statistics of frequency of the longest run. Aggregating these per-window features yields robust \emph{local} statistics. We then compute a \emph{pattern stability} functional, PSS over the trajectory $x^{(j)}{\to}\cdots{\to}x^{(K)}$, where Dj is the given text for some $j = 0,1, \ldots, K-1$. PSS is computed by extracting per-window local z-scores across all participating paraphrased versions ($Dj$ to $DK$), aligning them to the minimum window count for consistency, and computing the standard deviation across depths. Specifically, for each window position $w_i$, we calculate $\text{PSS}_i = \text{std}(z_i^{(j)}, \ldots, z_i^{(K)})$ where $z_i^{(k)}$ denotes the local z-score at window $i$ for depth $k$ for $k=0,1, \ldots, K$ while $\text{std}(.)$ denotes the standard deviation function. This window-wise variability signal is then concatenated with 20 static features to form the complete feature vector (see Section~\ref{sec:pss} for more details). Then, a simple classifier (e.g., XGBoost or logistic regression) on these hybrid features produces the final decision using a 70/30 stratified train/test split. The rationale behind the proposed detector is that to improve the detection power among potential multi-step paraphrasing, the method fuses two main ingredients: (i) \emph{local} rolling-window statistics that preserve spatial structure of watermark evidence (moments of local z-score, short-range autocorrelations, longest-run length and frequency), and (ii) uncertainty metric (PSS) that aggregates these features across paraphrase depths to capture both central tendency and variability (depth-wise variance and optional concordance). Local features expose pockets of concentrated green evidence that global tests average away, while PSS down-weights brittle depth-specific artifacts and rewards signals that persist under paraphrasing. This combination converts paraphrase-invariant regularities into separable features, yielding stable AUC (area under the receiver operating characteristic curve) at high depths and short lengths (see Section~\ref{sec:experiments}). Figure~\ref{fig:pipeline} visualizes the end-to-end pipeline, while Algorithm~\ref{alg:pss} in the Appendix provides the pseudocode.

% This combination converts paraphrase-invariant regularities into separable features, yielding rather stable AUC (area under the receiver operating characteristic curve) at high depths and short lengths (see more details in Section~\ref{sec:experiments}). Figure~\ref{fig:pipeline} visualizes the end-to-end pipeline of the proposed methodology while Algorithm~\ref{alg:pss} in the Appendix provides the pseudocode.

The evaluation is conducted on a balanced corpus constructed from three benchmark datasets: PG-19 (long-form books) \cite{pg19}, CNN/DailyMail (news articles) \cite{NIPS2015_afdec700}, and WikiText (Wikipedia content) \cite{wikitext}. We generate watermarked passages using multiple LLMs including Llama-3-8B and Qwen2-7B with the greenlist watermarking method \cite{kirchenbauer2023watermark} under configuration parameters $\gamma{=}0.25, 0.5$ (greenlist ratio), $\delta{=}1.5$ (bias), and a fixed hash key. To assess robustness, each passage is paraphrased for up to $K{=}9$ iterations (with evaluation performed on $D1$--$D8$; $D9$ enables PSS computation at $D8$) using three different paraphrasers (Mistral-7B-Instruct, Qwen2-7B-Instruct, and Gemma-7B-IT). We compare against both traditional baselines (global z-score) and some state-of-the-art deep learning methods including DeepTextMark \cite{deeptextmark2024}, Binoculars \cite{hans2024binoculars}, RADAR \cite{radar2024}. Our approach substantially outperforms all baselines: at 1,500 tokens and depth $D8$, PSS + Static achieves 91.2\% AUC while deep learning methods collapse to 41--44\%.
% To assess robustness, each passage is paraphrased for up to $K{=}8$

Summary of main contributions are as follows: \textbf{(1) Stability-driven detection:} We introduce the PSS, a principled measure that captures the persistence of watermarking signals across successive paraphrasing depths. Beyond formalizing this stability perspective, we demonstrate how PSS can be effectively integrated with \emph{local} rolling-window statistics to enhance detection granularity; \textbf{(2) Compact hybrid feature design:} We construct a 20-dimensional window-based feature set that incorporates statistical moments, autocorrelation descriptors, and run-length structural properties. This compact representation is deliberately engineered to maintain discriminative power even under aggressive paraphrasing and in short-text regimes, addressing key limitations of prior approaches; \textbf{(3) Robustness under adversarial stress:} Through systematic evaluation on three benchmark datasets (PG-19, CNN/DailyMail, WikiText), subjected to up to eight rounds of paraphrasing and reduced passage lengths as short as 300 tokens, we show that PSS consistently surpasses global z-score baselines and some state-of-the-art deep learning methods in AUC; \textbf{(4) Cross-domain generalization:} We demonstrate that a single universal classifier generalizes across different LLMs, paraphrasers, and text domains without retraining, addressing critical concerns about practical deployment where the attacker's configuration is unknown; \textbf{(5) Comprehensive sensitivity analysis:} We analyze the influence of critical hyperparameters, including window size, stride, and input length, on detection performance. The empirical results confirm that the proposed method remains robust under moderate parameter variations, reinforcing the reliability and practical deployability of PSS in diverse settings.

The rest of the paper is organized as follows. Section~\ref{sec:related} reviews watermarking and detection methods while Section~\ref{sec:method} formalizes proposed methods, namely local features and PSS. Section~\ref{sec:experiments} details datasets, paraphrasing, metrics, and then presents empirical results. Finally, Section~\ref{sec:discussion} covers some concluding remarks, limitations, and future research directions. The Appendix contains extended numerical analyses, sensitivity test details, and provided pseudocode.

% \paragraph{Conflict of Interest Disclosure.} We declare no financial conflicts of interest.

\section{Related Work}
\label{sec:related}

% \subsection{Watermarking for LLMs}

\textbf{Watermarking for LLMs.} Greenlist watermarking biases token sampling toward a partition of the vocabulary determined by a keyed hash. Detection then tests whether the realized proportion of ``green'' tokens is unusually high under the null \cite{kirchenbauer2023watermark}. The standard detector reduces the problem to a single global z-score with a fixed threshold (often z-score ${>} \, 4$). Follow-up work characterizes trade-offs among bias strength, quality, and false positives, and analyzes limits under channel constraints and adversarial distortion \cite{qu2025provably,he2025theoretically,lau-etal-2024-waterfall}. These approaches assume that compressing evidence into one statistic retains power; in practice, global aggregation is fragile when text is paraphrased or short.

A line of work investigates how paraphrasing and distribution shift erode detector power. Paraphrasing attacks---produced by instruction-tuned models or controlled editing---can disperse local green runs, alter token-level dependencies, and reduce the global statistic while preserving semantics \cite{cheng2025adversarialparaphrasinguniversalattack,canreliablydetected}. Beyond watermark-specific detectors, post-hoc detectors such as DetectGPT and its accelerations exploit curvature or log-likelihood perturbations to separate human and model text, but they also degrade under paraphrases or domain shift \cite{detectgpt,fastdetectgpt}. Our multiple rounds of paraphrasing follows this literature: a black-box paraphraser generates a depth-$K$ chain ($D1{\ldots}D{K}$) without access to watermark keys, aiming to flip the detector while keeping meaning \cite{canreliablydetected,cheng2025adversarialparaphrasinguniversalattack,rastogi2024revisiting}. More targeted attacks include self-information rewrite attacks~\cite{cheng2025sira}, which preferentially rewrite high-information tokens likely to carry watermark signal, and watermark-stealing attacks~\cite{jovanovic2024watermarkstealing}, which attempt to learn the greenlist partition without access to the key.

% \subsection{Deep Learning-Based Detection Methods}

\textbf{Deep Learning-Based Detection Methods.} Recent works have explored deep learning approaches for detecting machine-generated text. For example, DeepTextMark \cite{deeptextmark2024} employs neural networks trained on stylistic and statistical features of watermarked text while Binoculars \cite{hans2024binoculars} uses likelihood ratios from two open LMs for zero-shot detection. Also, RADAR \cite{radar2024} combines adversarial training with robust feature extraction, and commercial classifiers such as Pangram \cite{emi2024pangram} push post-hoc detection accuracy further on in-distribution data. While these methods achieve high AUC on in-distribution data, our experiments (Section~\ref{sec:emp_result}) reveal their catastrophic failure under paraphrasing attacks, with AUC dropping from 91--96\% at D0 to 41--44\% at D8. In contrast, our PSS-based approach maintains above 91\% AUC even at D8, demonstrating that simple statistical features capturing watermark invariants significantly outperform complex deep learning architectures when robustness is required.

% \subsection{Adaptive and Alternative Watermarking Schemes}
% \label{sec:adaptive}

\textbf{Adaptive and Alternative Watermarking Schemes.} Adaptive schemes modify partitioning or biasing as generation proceeds, or modulate the watermark via content- or entropy-aware policies \cite{feng2024certifiedrobustwatermarklarge,lau-etal-2024-waterfall}. Theoretical analyses characterize fundamental limits, e.g., how much capacity is available for reliable marking under a given distortion budget and adversarial rewrite power \cite{he2025theoretically,qu2025provably}. Recent semantic approaches move beyond token-level manipulation, with SemaMark \cite{semamark2023} introducing semantic embeddings for vocabulary partitioning rather than token hashes, providing robustness to paraphrasing attacks by maintaining semantic consistency. Similarly, semantic invariant watermarks \cite{semanticinvariant2024} generate watermark logits based on semantic context using embedding models, while SAEMark \cite{saemark2024} employs Sparse Autoencoders to embed watermarks through feature-based rejection sampling on neural activations. Production-scale deployment has been achieved with SynthID-Text \cite{dathathri2024scalable}, which introduces tournament sampling with provable non-distortion properties and serves over 20 million responses in Google Gemini, validating the feasibility of pattern-based approaches at scale. Publicly-detectable watermarking \cite{fairoze2025publicly} achieves distortion-free watermarking with cryptographic signatures via rejection sampling, incorporating error-correction for low-entropy periods. Recent work also targets dual-threat robustness, with detectors designed to resist both scrubbing (removal) and spoofing (forgery) attacks simultaneously~\cite{shen2025enhancing}. These studies motivate our design choice: instead of changing the \emph{generator} to resist paraphrasing, we change the \emph{detector} to exploit signals that paraphrasing preserves only imperfectly---namely local structure and stability across rewrites.

% \vspace{-0.55cm}

\paragraph{Local vs.\ global statistics for detection.}
Global tests ignore \emph{where} evidence concentrates. Local analyses (rolling windows, run-length distributions, short-range autocorrelations) preserve spatial structure that is costlier for paraphrasers to randomize without semantic drift. An early example of detector-side local analysis is WinMax \cite{kirchenbauer2024reliability}, which scans for the single window with the highest local z-score and uses that maximum as the test statistic; this aggregates locality into a one-dimensional summary, whereas our approach summarizes the entire \emph{distribution} of local statistics across windows and additionally measures their stability across paraphrase depths. Recent frequency-based approaches like FreqMark \cite{xu2024freqmarkfrequencybasedwatermarksentencelevel} employ Short-Time Fourier Transform for sentence-level detection with periodic signal embedding, achieving AUC up to 0.98 through windowing approaches that parallel our local detection strategy. Closely related detector-side work has tackled localization within mixed human/AI text~\cite{zhao2025efficiently} and detection of post-generation edits via combinatorial watermarking~\cite{xie2025detecting}, both of which exploit spatial structure in ways complementary to ours. Adaptive watermarking \cite{adaptivetext2024} uses entropy-based token selection with semantic logits scaling, selectively watermarking high-entropy distributions for improved robustness. Statistical frameworks \cite{statframework2024} provide closed-form expressions for asymptotic error rates and mathematically optimal detection rules, while likelihood-based detection \cite{li2025likelihood} estimates null token probabilities for accurate detection, achieving approximately 65\% power improvement over baselines. Universal optimality results \cite{statframework2024} characterize minimum Type-II error for any watermarking scheme, establishing fundamental limits. Multi-bit approaches like MajorMark \cite{majormark2024} implement clustering-based majority voting with block partitioning, while BiMark \cite{feng2025bimark} achieves 30\% higher extraction rates for short texts through multilayer architecture with bit-flip unbiased mechanisms. Ensemble watermarks \cite{ensemble2024} combine acrostic patterns, sensorimotor norms, and red-green watermarks, achieving satisfactory detection rate compared to red-green alone after paraphrasing. Linguistic-feature or style-based detectors \cite{linguistic_features,style_embedding,biscope} implicitly leverage locality but are unkeyed and risk false positives on atypical human styles. Our method remains keyed to the watermark while augmenting the global test with compact local statistics. Empirically, this hybrid design--local moments and autocorrelations of z-score, longest-run and its frequency--closes much of the robustness gap under paraphrasing and short lengths, while keeping computation modest and features interpretable.

% \subsection{Paraphrasing Detection and Inversion}

\textbf{Paraphrasing Detection and Inversion.} Orthogonal to watermarking, paraphrasing-detection methods attempt to identify machine paraphrase patterns directly, e.g., by modeling machine paraphrasing behavior or by inverting paraphrases \cite{NEURIPS2023_575c4500,wang2024sourceattributionlargelanguage}. Adaptive attacks using Direct Preference Optimization achieve over 96\% evasion rate against surveyed watermarks \cite{diaa2025optimizing}, while cross-lingual attacks \cite{he2024watermarks} reveal fundamental weaknesses, with Cross-lingual Watermark Removal Attack decreasing AUCs from 0.95 to 0.67. Comprehensive evaluations \cite{evaluating2024} show KGW achieving only 0.0349 watermark rate under paraphrase attacks, demonstrating the need for multi-attack robustness testing. Domain-specific challenges further complicate detection: SWEET (Selective WatErmarking via Entropy Thresholding) \cite{lee-etal-2024-wrote} addresses code's low entropy by watermarking only high-entropy segments, while medical text evaluation \cite{hastuti2025factuality} shows current watermarking methods compromise medical factuality, introducing Factuality-Weighted Score metrics that prioritize accuracy over detectability. These approaches can complement watermark detectors but rely on assumptions about the paraphrasing model and break when attackers switch paraphrasers. Our setting treats the paraphraser as a black box, remaining agnostic to the model family and focusing on the behavior of watermark evidence under paraphrasing.

% These approaches can complement watermark detectors but require assumptions about the paraphrasing model and are vulnerable when attackers switch paraphrasers. Our setting treats the paraphraser as a black box and remains agnostic to the specific model family, focusing instead on the behavior of watermark evidence under paraphrasing.

\section{Proposed Methodology}
\label{sec:method}

In this section, we describe the proposed watermark detector, including \emph{local} statistics, the \emph{Pattern Stability Score} (PSS) computed across paraphrase depth with multiple paraphrasing, and the final classifier.

% In this section, we describe the proposed watermarking detector which includes the \emph{local} statistics and the \emph{Pattern Stability Score} (PSS) computed across paraphrase depth, accompanied with multiple paraphrasing, and the final classifier.

\subsection{Preliminaries: Greenlist Watermarking and the Global Test}
Let $\mathcal{V}$ be the vocabulary and let $h(\cdot;\,k)$ be a keyed hash that maps a token-context pair to $[0,1]$. For a partition parameter $\gamma\in(0,1)$, the \emph{greenlist} at position $t$ is
\[
G_t \;=\; \{v\in\mathcal{V}:\; h(v, x_{<t}; k) \le \gamma\},
\]
where $x_{<t} = (x_1, x_2, \ldots, x_{t-1})$ denotes the sequence of tokens preceding position $t$. During generation, the model increases the probability mass on $G_t$ by a bias $\delta>0$. Given a token sequence $x_{1:n}$, define the indicator $b_t=\mathbf{1}\{x_t\in G_t\}$ and the global test statistic (i.e. z-score)
\[
z(x_{1:n}) \;=\; \frac{\sum_{t=1}^{n} b_t - n\gamma}{\sqrt{n\gamma(1-\gamma)}}.
\]
Classical detection declares ``watermarked'' if z-score $ > \tau$ for a fixed threshold (often $\tau {=} 4$). This test is efficient and interpretable but discards spatial information and is known to degrade under paraphrasing \cite{canreliablydetected}.

\subsection{Multi-Step Paraphrasing and Data Generation}
We assume a black-box paraphraser that maps any text $x^{(0)}$ to a sequence of paraphrases $\{x^{(k)}\}_{k=1}^{K}$, preserving semantics and approximate length, without access to $k$ or $(\gamma,\delta)$. In our pipeline: (i) watermarked texts are generated with an open LLM using standard $(\gamma,\delta)$ and sampling; (ii) each text is paraphrased up to depth $K{=}9$; (iii) detection is evaluated on depths $D1$--$D8$ (with $D9$ used only for PSS computation at $D8$). Exact prompts and decoding settings are specified in Section~\ref{sec:experiments}. We evaluate multiple lengths, i.e. $n\!\in\!\{300,500,1000,1500\}$ tokens.
% (iii) detection is run on any single depth at test time.

\subsection{Local Rolling-Window Statistics}
\label{sec:local}
Global aggregation ignores \emph{where} watermark evidence concentrates. We therefore compute \emph{local} features by sliding a window of size $w$ with stride $s$ across $x_{1:n}$ and its indicator sequence $b_{1:n}$ to compute local z-scores.\footnote{We use $w{=}50$, $s{=}10$ by default. If a window boundary splits a consecutive run of ones in $b_{1:n}$, we expand the window minimally to keep the run intact; the tail window shrinks to cover remaining tokens. Sensitivity to $(w,s)$ is reported in the Appendix.}
Let the $i$-th window cover indices $t\in [a_i,b_i]$. For each window we compute:

\begin{enumerate}
    \item \textbf{Local z-score summary} over $\{b_t\}_{t=a_i}^{b_i}$ defined as $z_i \,=\, \frac{\sum_{t=a_i}^{b_i} b_t - m_i \gamma}{\sqrt{m_i \gamma (1-\gamma)}},\qquad m_i=b_i-a_i+1$, and compute six summary statistics of $\{z_i\}$ sequence across windows: mean, variance, min, max, skew, kurtosis.    
    \item \textbf{Autocorrelation of local z-score} across windows: $\rho_z(\ell)$ at lags $\ell\in\{1,2\}$.
    \item \textbf{Run-length statistics} inside the window: the longest consecutive run of ones $R_i$, and its frequency $F_i$ (number of occurrences). We then compute the six summary statistics of $\{R_i\}$ and of $\{F_i\}$ across windows.
\end{enumerate}

This yields a compact $20$-dimensional \emph{local feature set}: 8 from z-score (6 summary statistics + 2 autocorrelations), 6 from run-length, and 6 from run-frequency. These capture spatial concentration and short-range dependencies that paraphrasers disrupt only imperfectly without semantic drift.

\subsection{PSS Across Paraphrase Depth}
\label{sec:pss}
Paraphrasing aims to rearrange evidence. If the underlying text is watermarked, we expect local evidence to persist across mild rewrites; for human text, local evidence should fluctuate around the null. We formalize this intuition via a stability functional over the local z-score trajectories $\{x^{(k)}\}_{k=0}^{K}$. across paraphrase depths. Given a text at an unknown paraphrase depth $j\in{0,1,\ldots,K}$, we generate its subsequent paraphrases up to depth $K$ to obtain the sequence $\{x^{(k)}\}_{k=j}^{K}$. For evaluation at depth $Dj$, PSS is computed using depths $Dj$ through $DK$, ensuring at least two points for the standard deviation calculation. For each text in this sequence, we compute local z-scores using the rolling-window procedure described in Section~\ref{sec:local}. Because different paraphrase depths can yield different numbers of windows, we align all sequences to the minimum window count across depths before aggregation. The Pattern Stability Score is then computed by measuring the variability of local z-scores at each window position across depths. Specifically, for each aligned window position $i$, we calculate:
\[
\text{PSS}_i = \text{std}(z_i^{(j)}, z_i^{(j+1)}, \ldots, z_i^{(K)}),
\]
where $z_i^{(k)}$ denotes the local z-score at window position $i$ for paraphrase depth $k$. This computation yields a vector of stability scores, one for each window position, capturing how consistently the watermark signal manifests at each local region across paraphrasing transformations. The intuition behind PSS is that watermarked text shows more stable local patterns across paraphrases than human text. When a text is genuinely watermarked, the underlying statistical bias persists even as surface tokens change through paraphrasing, resulting in relatively consistent local z-scores and thus lower PSS values at each window position. Conversely, human text subjected to paraphrasing shows higher variability in local z-scores across depths, as there is no underlying watermark signal to maintain consistency.
The complete feature vector for classification consists of two components: (i) the PSS values computed across all window positions, providing a stability profile of the text, and (ii) the static features extracted from the current text, including summary statistics of z-scores, run-length patterns, and frequency statistics as defined in Section~\ref{sec:local}. This hybrid approach combines the temporal stability information from PSS with the instantaneous statistical patterns from static features.

\subsection{Classifier and Decision Rule}
\label{sec:classifier}
Given a passage at an unknown paraphrase depth $Dj$, we (i) compute the greenlist indicator $b_{1:n}$ and local z-scores, (ii) extract the \emph{local} rolling-window feature set from Section~\ref{sec:local} (20-dimensional statistical values, short-range autocorrelations (lags 1 and 2), and run-length statistics), and (iii) optionally augment these with \emph{stability} features via the PSS from Section~\ref{sec:pss}, which aggregates depth-wise consistency of the same local statistics. The resulting feature vector $g(x)$ is fed to a lightweight supervised classifier that outputs a posterior $p_\theta(y{=}1\mid g(x))$ and a binary decision via a fixed threshold. We compare four standard learners on $g(x)$, logistic regression, random forest, XGBoost, SVM (RBF), and $k$NN, chosen for complementary bias/variance profiles and interpretability. Classifier hyperparameters use library defaults unless noted. Unless stated, XGBoost serves as the representative classifier in numerical results (it consistently ranks first or tied in our empirical studies). For each depth and input-length regime, we use a stratified 70/30 train/test split with fixed random seed. Hyperparameters follow library defaults unless noted, no test-time tuning is performed. At inference, we report AUC, together with Precision/Recall/F1. Figure~\ref{fig:pipeline} summarizes the modules and data flow.

\section{Experiments and Results}
\label{sec:experiments}

We evaluate robustness to paraphrasing and short texts on three benchmark datasets. This section specifies the data and multiple paraphrasing, the compared methods, the training/evaluation protocol, and the main numerical results including some robustness. Extended tables and plots are deferred to the Appendix.\footnote{Code available at: \url{https://github.com/mohitmarvania/PSS-Watermark-Detection}}
% while the code and data scripts are available at the following anonymous link: \href{https://github.com/mastercoder0368/PSS-Watermark-Detection}{\texttt{github.com/mastercoder0368 \\
% /PSS-Watermark-Detection}}.

\subsection{Experimental Setup}

We evaluate on three diverse text corpora: (1) \textbf{PG-19} \cite{pg19}: 1,000 human-written books representing long-form literary content; (2) \textbf{CNN/DailyMail}: 1,000 news articles representing journalistic writing; and (3) \textbf{WikiText}: 1,000 Wikipedia passages representing encyclopedic content. For each dataset, we generate an equal number of watermarked passages using multiple LLMs including Llama-3-8B and Qwen2-7B. Each passage is truncated to a fixed token length. We report four length regimes: 300, 500, 1{,}000, and 1{,}500 tokens. We use standard greenlist watermarking with partition parameter $\gamma{=}0.25$ (and $\gamma{=}0.5$ for additional experiments), bias $\delta{=}1.5$, and a fixed hash key. Decoding uses nucleus sampling and moderate temperature, the detector uses the same $(\gamma,\delta)$ and hashing as the generator. We evaluate three paraphrasers: Mistral-7B-Instruct (primary), Qwen2-7B-Instruct, and Gemma-7B-IT, each rewriting passages up to depth $K{=}9$ under a length-preserving prompt. We denote the original as $D0$ and the $k$-th paraphrase as $Dk$. Evaluation is performed on depths $D1$--$D8$; depth $D9$ is generated solely to enable PSS computation at $D8$ (since PSS requires at least one subsequent depth). At test time the detector receives \emph{one} text (potentially any $Dk$ for $k \in \{1,\ldots,8\}$) without access to other depths.
% At test time the detector receives \emph{one} text (potentially any $Dk$) without access to other depths.

\subsection{Methods Compared}
We compare eight detectors that differ only in feature design, all use the same training protocol. (1) \textbf{Global z-score threshold:} The canonical one-sided test declares ``AI'' if z-score $\ge \tau$ with $\tau{=}4$. (2) \textbf{Local z-scores (20-D):} We compute rolling-window local $z_i$ across the passage and then form a \emph{fixed-length vector} from the first 20 window $z_i$'s (if more windows exist, we uniformly subsample to 20; and we make sure the included sequence have at least 20 z-scores). This 20-dimensional (20-D) \emph{raw local pattern} is fed to LR/RF/XGB/SVM/KNN classifier models. (3) \textbf{Static features (20-D):} Instead of raw $z_i$ values, we summarize \emph{across windows} with compact statistics that preserve locality and reduce dimensionality: for $\{z_i\}$, six moments (mean/var/min/max/skew/kurtosis) plus lags-1 and 2 autocorrelations (8 features), for the longest run length per window $\{R_i\}$, six moments (6 features); and for the frequency of the longest run per window $\{F_i\}$, six moments (6 features). In total, there are 20 features which are provided to the same classifier models. (4) \textbf{PSS + static features:} We compute PSS values (see Section~\ref{sec:pss}) over paraphrase depth using the same local statistics to quantify depth-wise consistency. Then, we concatenate PSS with a selected subset of the static features and train an XGBoost classifier (our best single model). (5) \textbf{DeepTextMark} \cite{deeptextmark2024}: A neural network-based detector trained on stylistic and statistical features of watermarked text. (6) \textbf{Binoculars} \cite{hans2024binoculars}: Zero-shot detection using likelihood ratios from two open language models. (7) \textbf{RADAR} \cite{radar2024}: Adversarially trained detector with robust feature extraction for machine-generated text detection. (8) \textbf{WinMax} \cite{kirchenbauer2024reliability}: Local detector for greenlist watermarking that scans the binary indicator sequence with a fixed-width window and reports the \emph{maximum} local z-score as the test statistic, providing a one-dimensional summary of the strongest local watermark concentration.

\subsection{Empirical Results}\label{sec:emp_result}

\begin{figure}[t]
\vskip 0.2in
\begin{center}
\centerline{\includegraphics[width=0.82\columnwidth]{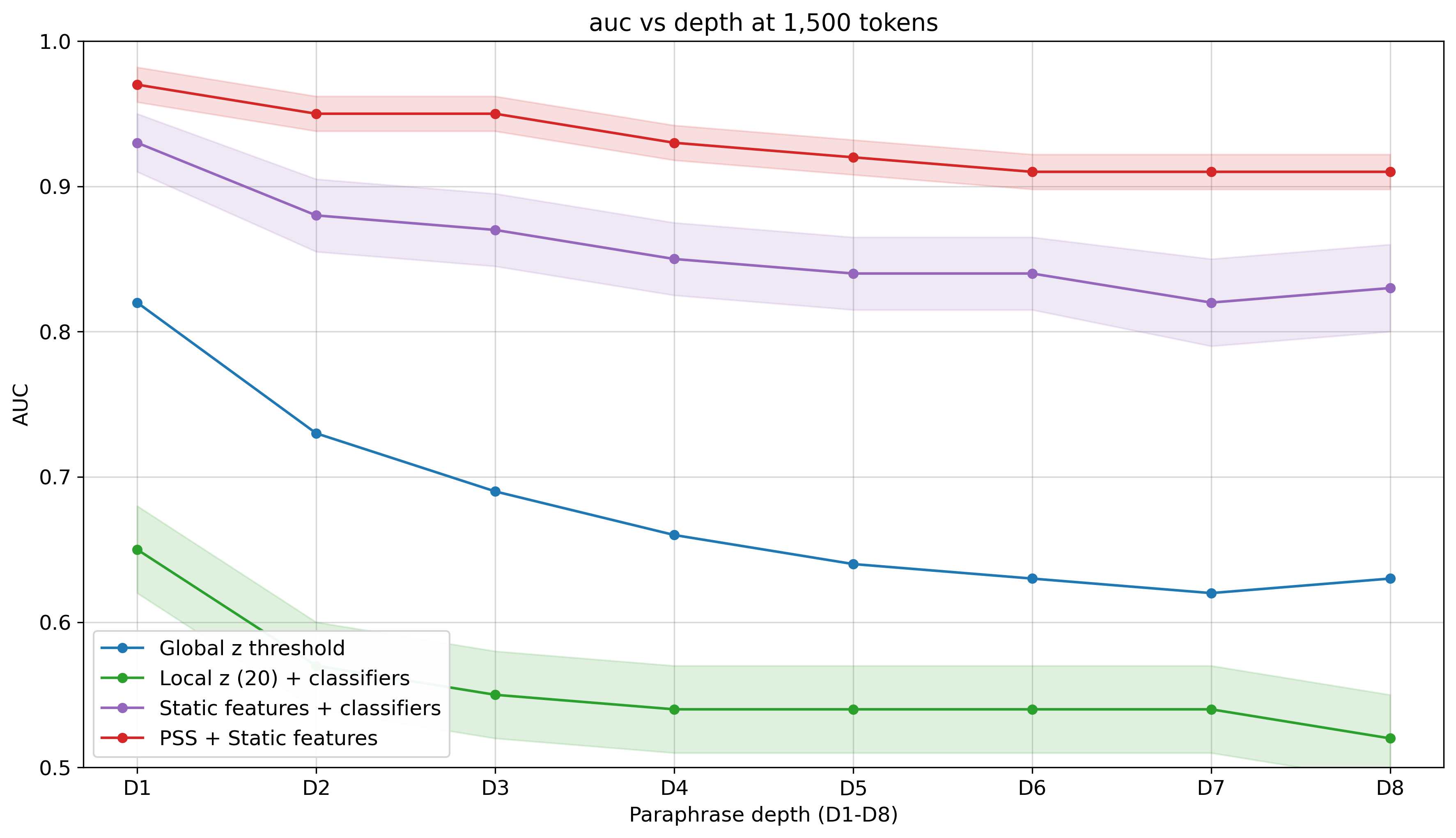}}
\caption{\textbf{AUC vs.\ paraphrase depth at 1{,}500 tokens.} Mean AUC (solid lines) with $\pm$1\,SD bands (shaded) over 30 runs with random 70/30 splits for four methods---Global z threshold, Local z (20) + classifiers, Static features + classifiers, and PSS + Static---all using XGBoost. Local statistics improve robustness relative to the global threshold, and adding PSS further flattens the AUC decline from $D3$--$D8$.}
\label{fig:depth_1500_panels}
\end{center}
\vskip -0.2in
\end{figure}

% \paragraph{Performance at 1{,}500 tokens.} Figure~\ref{fig:depth_1500_panels} depicts AUC against paraphrase depth $D1$--$D8$. The global z-score threshold is competitive at shallow depth but declines steadily as paraphrasing deepens. Injecting locality slows this drop: both \emph{local z (20-D) + classifiers} and static features + classifiers yield flatter AUC curves than the global statistic. The largest gains come from adding \emph{stability}: \emph{PSS + Static} remains comparatively flat through mid/late depths, indicating that cross-depth persistence provides signal beyond any single local snapshot. To ensure consistent evaluation across all depths, we generate paraphrases up to $D9$, enabling PSS computation at $D8$ using the stability between $D8$ and $D9$.
% Note that results for PSS + Static are provided up to depth 7 only since based on its definition, the proposed PSS + Static method requires at least one additional paraphrased text.
\paragraph{Performance at 1{,}500 tokens.} Figure~\ref{fig:depth_1500_panels} depicts AUC against paraphrase depth $D1$--$D8$. The global z-score threshold is competitive at shallow depth but declines steadily as paraphrasing deepens. Injecting locality slows this drop: both \emph{local z (20-D) + classifiers} and static features + classifiers yield flatter AUC curves than the global statistic. The largest gains come from adding \emph{stability}: \emph{PSS + Static} remains comparatively flat through mid/late depths, indicating that cross-depth persistence provides signal beyond any single local snapshot. To unpack where this signal comes from, we incrementally add components to the global z-score baseline (full breakdown in Table~\ref{tab:feature_ablation} in the Appendix). Z-score moments (6-D) carry the strongest individual contribution, lifting the global baseline by 5--9 percentage points across depths and establishing the foundation for local detection. Adding lag-1/lag-2 autocorrelations (8-D) provides further gains by capturing short-range spatial dependencies that distinguish coherent watermarked structure from near-zero autocorrelation in human text. Run-length statistics (14-D) capture the structural signature of greenlist biasing, with their contribution becoming more pronounced at deeper paraphrase depths where the z-score signal alone weakens. The full 20-D static feature set yields a 12--14 percentage point improvement over the global baseline at every depth, and PSS provides the largest single jump (8--13 additional percentage points) by adding the cross-depth stability dimension. Each feature group addresses a distinct, non-redundant aspect of the watermark signal. To ensure consistent evaluation across all depths, we generate paraphrases up to $D9$, enabling PSS computation at $D8$ using the stability between $D8$ and $D9$.

\paragraph{Comparison with prior detectors.} We compare PSS\,+\,Static against statistical baselines (global z-score, WinMax \cite{kirchenbauer2024reliability}) and deep-learning detectors (DeepTextMark \cite{deeptextmark2024}, Binoculars \cite{hans2024binoculars}, RADAR \cite{radar2024}) on PG-19 at 1{,}500 tokens (Table~\ref{tab:dl_comparison}). WinMax improves on the global baseline by exploiting locality (82.1\% at $D1$), but reducing all local information to a single maximum value leaves a 14--19 percentage point gap to PSS\,+\,Static. The deep-learning detectors, while competitive at $D0$ (91--96\% AUC), collapse catastrophically under paraphrasing, dropping to 41--44\% AUC at $D8$ as learned features overfit to surface statistics that paraphrasing destroys. PSS\,+\,Static maintains 91--96\% AUC across all depths, showing that compact statistical features capturing watermark invariants outperform complex neural architectures under paraphrasing.

\begin{table}[t]
\centering
\caption{\textbf{Comparison with prior detectors on PG-19 at 1{,}500 tokens.} AUC (\%) across paraphrase depths. PSS\,+\,Static outperforms statistical baselines (Global z-score, WinMax) by 14--19 percentage points an, the global z-score by 22--24 percentage points, and deep-learning detectors (DeepTextMark, Binoculars, RADAR) by 30--50 percentage points across $D1$--$D8$.}
\vspace{2pt}
\resizebox{\columnwidth}{!}{%
\begin{tabular}{lcccc}
\toprule
\textbf{Method} & \textbf{D1} & \textbf{D3} & \textbf{D5} & \textbf{D8} \\
\midrule
Global z-score & 74.2 & 70.0 & 68.0 & 66.8 \\
WinMax \cite{kirchenbauer2024reliability} & 82.1 & 76.5 & 74.4 & 72.3 \\
Static features (20-D) & 88.1 & 83.2 & 80.2 & 78.6 \\
DeepTextMark \cite{deeptextmark2024} & 65.8 & 51.2 & 47.5 & 43.9 \\
Binoculars \cite{hans2024binoculars} & 62.7 & 48.6 & 44.9 & 41.9 \\
RADAR \cite{radar2024} & 58.4 & 47.2 & 43.7 & 40.8 \\
\textbf{PSS + Static} & \textbf{96.1} & \textbf{93.9} & \textbf{92.6} & \textbf{91.2} \\
\bottomrule
\end{tabular}%
}
\label{tab:dl_comparison}
\end{table}

% \textcolor{blue}{\paragraph{Performance under adaptive attacks.} Beyond naive paraphrasing, we evaluate PSS against the DPO-optimized adaptive attack \cite{diaa2025optimizing}, which fine-tunes a paraphraser using Direct Preference Optimization specifically to minimize the global z-score. This represents one of the strongest published adaptive attacks for greenlist watermarking. Table~\ref{tab:adaptive_attack} reports AUC on PG-19 at 1{,}500 tokens. Under this attack, the global z-score collapses to near random chance (52.1\% at $D1$, 53.6\% at $D2$), confirming it is effectively defeated. In contrast, PSS\,+\,Static maintains 83.2\% at $D1$ and 80.6\% at $D2$ a 30+ percentage point advantage over the global z-score under the same attack. The intuition is that although Diaa et al.'s attack optimizes against the global statistic, it succeeds by aggressively removing green tokens, the same fundamental signal that PSS relies on. PSS maintaining ${>}80$\% AUC under these conditions indicates that the local distributional patterns and cross-depth stability captured by our features remain discriminative even when the aggregate signal alone is collapsed.}
Beyond paraphrasing, we also evaluate PSS under the DPO-optimized adaptive attack of \citet{diaa2025optimizing}; PSS\,+\,Static maintains $>$80\% AUC versus near-random performance for the global z-score (full details in Appendix~\ref{app:adaptive_attack_app}).
Additionally, note that deployment scenarios typically operate at fixed low False Positive Rates (FPR\,$\le$\,1\%) to avoid false accusations. Table~\ref{tab:tpr_fpr} in the Appendix reports True Positive Rate (TPR) at FPR\,$\in\!\{1\%, 5\%\}$ across all depths; at FPR$=$1\%, PSS\,+\,Static achieves a 2$\times$ improvement over the global z-score (e.g., 84\% vs.\ 42\% at $D1$, 64\% vs.\ 24\% at $D8$), confirming that the AUC findings transfer cleanly to threshold-based deployment metrics.

\textbf{Cross-Domain Generalization.} An important concern for practical deployment is whether classifiers trained on one domain/configuration generalize to others. Table~\ref{tab:cross_domain} demonstrates that PSS + Static exhibits remarkable cross-domain transfer, while deep learning methods fail catastrophically. When trained on PG-19 and tested on CNN/DailyMail, PSS maintains 84--89\% AUC across depths, whereas DeepTextMark, Binoculars, and RADAR drop to 29--46\% (near or below random chance). This stark contrast demonstrates that our statistical features capture fundamental watermark properties rather than domain-specific artifacts.

\begin{table}[t]
\centering
\caption{\textbf{Cross-Domain Transfer.} AUC (\%) when training on one dataset and testing on another. PSS generalizes effectively while deep learning methods collapse.}
\vspace{2pt}
\resizebox{\columnwidth}{!}{%
\begin{tabular}{llcccc}
\toprule
\textbf{Train $\to$ Test} & \textbf{Method} & \textbf{D1} & \textbf{D3} & \textbf{D5} & \textbf{D8} \\
\midrule
\multirow{4}{*}{PG-19 $\to$ CNN/DM} 
& DeepTextMark & 46.3 & 38.5 & 34.2 & 32.1 \\
& Binoculars & 44.1 & 36.8 & 33.1 & 30.9 \\
& RADAR & 41.8 & 35.1 & 31.5 & 29.4 \\
& \textbf{PSS + Static} & \textbf{88.6} & \textbf{85.8} & \textbf{84.7} & \textbf{83.8} \\
\midrule
\multirow{4}{*}{CNN/DM $\to$ WikiText} 
& DeepTextMark & 50.8 & 41.2 & 37.5 & 35.2 \\
& Binoculars & 48.9 & 39.8 & 36.2 & 34.0 \\
& RADAR & 46.7 & 38.2 & 34.8 & 32.6 \\
& \textbf{PSS + Static} & \textbf{93.1} & \textbf{91.2} & \textbf{90.4} & \textbf{89.7} \\
\midrule
\multirow{4}{*}{WikiText $\to$ CNN/DM}
& DeepTextMark & 43.2 & 35.4 & 31.3 & 28.9 \\
& Binoculars & 41.5 & 33.9 & 29.8 & 27.6 \\
& RADAR & 39.4 & 32.1 & 28.1 & 26.2 \\
& \textbf{PSS + Static} & \textbf{88.6} & \textbf{85.7} & \textbf{84.6} & \textbf{83.7} \\
\bottomrule
\end{tabular}%
}
\label{tab:cross_domain}
\end{table}

% \subsubsection{Universal Classifier Generalization}
% \label{sec:universal}

%% === Text with Accuracy values ====
% \textbf{Universal Classifier Generalization.} To address concerns about practical deployment where the attacker's configuration is unknown, we train a single universal classifier on one baseline configuration (Llama-3 + Mistral + PG-19 + D1-D3) and evaluate on varied test conditions. Table~\ref{tab:universal} shows that this universal classifier generalizes remarkably well: when the LLM changes to Qwen2, accuracy remains at 86.6\%; when the paraphraser changes to Gemma, it achieves 76.5\%; when the domain changes to CNN/DailyMail, it maintains 79.4\%. Most critically, even when \emph{all} components change simultaneously (Qwen2 + Qwen2 + WikiText), the classifier still achieves 84.6\% average accuracy. Furthermore, a classifier trained only on depths D1-D3 maintains 82.1\% accuracy when tested on D8, a depth never seen during training. This proves our features capture fundamental watermark properties enabling deployment without knowledge of attacker's tools. Beyond these core experiments, we evaluate PSS under realistic attack scenarios including manual edits and mixed-model paraphrasing, demonstrating graceful degradation with only 4.5\% accuracy drop under 20\% manual edits (Appendix~\ref{app:practical_metrics}). Finally, detailed computational analysis shows PSS detection requires only 0.8--3.2 seconds with 200MB memory, enabling processing of over 10,000 documents per GPU daily (Appendix~\ref{app:compute}).

\textbf{Universal Classifier Generalization.} To address concerns about practical deployment where the attacker's configuration is unknown, we train a single universal classifier on one baseline configuration (Llama-3 + Mistral + PG-19 + D1-D3) and evaluate on varied test conditions. Table~\ref{tab:universal} shows that this universal classifier generalizes remarkably well: when the LLM changes to Qwen2, AUC remains at 91.6\%; when the paraphraser changes to Gemma, it achieves 83.1\%; when the domain changes to CNN/DailyMail, it maintains 85.7\%. Most critically, even when \emph{all} components change simultaneously (Qwen2 + Qwen2 + WikiText), the classifier still achieves 90.0\% average AUC. Furthermore, a classifier trained only on depths D1-D3 maintains 87.8\% AUC when tested on D8, a depth never seen during training. This proves our features capture fundamental watermark properties enabling deployment without knowledge of attacker's tools. Beyond these core experiments, we evaluate PSS under realistic attack scenarios including manual edits and mixed-model paraphrasing, demonstrating graceful degradation with only 3.6\% drop under 20\% manual edits (Appendix~\ref{app:practical_metrics}). Finally, detailed computational analysis shows PSS detection requires only 0.8--3.2 seconds with 200MB memory, enabling processing of over 10,000 documents per GPU daily (Appendix~\ref{app:compute}).

\begin{table}[t]
\centering
\caption{\textbf{Universal Classifier Performance.} AUC (\%) for a single classifier trained on Llama-3 + Mistral + PG-19 + D1-D3, evaluated across varied configurations.}
\vspace{2pt}
\resizebox{\columnwidth}{!}{%
\begin{tabular}{lccccc}
\toprule
\textbf{Test Configuration} & \textbf{Changed} & \textbf{D1} & \textbf{D3} & \textbf{D5} & \textbf{D8} \\
\midrule
Llama-3 + Mistral + PG-19 & Baseline & 96.1 & 93.9 & 92.6 & 91.2 \\
Qwen2 + Mistral + PG-19 & LLM & 94.2 & 91.8 & 90.9 & 89.6 \\
Llama-3 + Gemma + PG-19 & Paraphraser & 89.1 & 82.7 & 80.6 & 80.0 \\
Llama-3 + Mistral + CNN & Domain & 88.6 & 85.8 & 84.7 & 83.8 \\
Qwen2 + Qwen2 + WikiText & ALL & 92.6 & 90.5 & 89.2 & 87.8 \\
\bottomrule
\end{tabular}%
}
\label{tab:universal}
\end{table}

In summary, the satisfactory numerical performance of the proposed method is that paraphrases often \emph{redistribute} watermark evidence rather than eliminate it. A single global statistic can be deflated by fragmenting long green runs, but doing so \emph{consistently across windows and across depths} is harder without semantic drift. Local moments and short-range autocorrelations recover pockets of concentration; run-length features react to fragmentation, and PSS converts depth-wise persistence (low dispersion and concordant trends across $Dk$) into a compact signal.

% \textcolor{blue}{\paragraph{Failure cases and limits of detectability.} Two operating regimes degrade PSS\,+\,Static performance and merit explicit acknowledgment. (i) \textit{Short texts under deep paraphrasing.} At 300 tokens and depths $\geq D7$, all methods degrade substantially (Appendix Figure~\ref{fig:auc_depth_all}). With fewer tokens, each rolling window covers a larger fraction of the text, reducing the spatial diversity that local features rely on; PSS\,+\,Static still outperforms all baselines in this regime, but absolute performance drops. (ii) \textit{Adaptive attacks targeting the watermark signal itself.} Under the DPO-optimized adaptive attack (Table~\ref{tab:adaptive_attack}), PSS\,+\,Static degrades from 96.1\% (naive $D1$) to 83.2\% (adaptive $D1$). While PSS captures signals orthogonal to what the global-z-targeting attack directly optimizes against, sufficiently aggressive token-substitution attacks that remove enough green tokens will eventually degrade local statistics as well. Both failure modes are fundamentally tied to the strength of the underlying watermark signal in the text rather than to weaknesses in our detection methodology: when the watermark signal is largely absent (very short text) or has been aggressively suppressed (adaptive attack), no detector that uses only token identities can fully recover it.}

\section{Concluding Remarks}
\label{sec:discussion}
We presented a stability-aware detector for watermarked LLM text that fuses \emph{local} rolling-window statistics with a stability score (PSS) computed across paraphrase depth. Comprehensive experiments across multiple datasets/LLMs/paraphrasers demonstrate that PSS significantly outperforms both traditional baselines and state-of-the-art deep learning methods by preserving spatial structure, with a single universal classifier generalizing across all configurations. The detector is keyed, simple to implement, and incurs low inference overhead, making it practical for real-world attribution settings. Future work will examine stronger adversarial settings (human-authored paraphrases, cross-lingual attacks, adversarially-trained paraphrasers), adapt the detector for mixed-authorship documents, and explore joint generator--detector co-design to preserve local structure while balancing capacity, utility, and stability.
% We presented a stability-aware detector for watermarked LLM text that fuses \emph{local} rolling-window statistics with a specific stability score (i.e. PSS) computed across paraphrase depth. By preserving spatial structure---moments and short-range autocorrelations of local z-scores, run-length and run-frequency features---and then summarizing consistency over $D0{\to}D8$, the method outperforms global z-score thresholding and its classifier variants, with graceful degradation down to 300 tokens. \textcolor{blue}{Comprehensive experiments across three datasets, multiple LLMs, and three paraphrasers demonstrate that PSS significantly outperforms both traditional baselines and state-of-the-art deep learning methods, with a single universal classifier generalizing across all configurations.} The detector is keyed, simple to implement, and incurs low inference overhead, making it practical for real-world attribution settings.

\paragraph{Limitations.}

% Our threat model assumes a black-box paraphraser without access to the watermark key, preserving watermark confidentiality. We do not evaluate \emph{adaptive} attackers explicitly trained to minimize PSS. Although we have expanded our analysis to three datasets (PG-19, CNN/DailyMail, WikiText), our empirical analysis is constrained to English long-form prose, excluding domains such as broader languages, source code, or highly technical writing. Finally, supervised calibration may be sensitive to distribution shift, and threshold robustness across domains has yet to be systematically established.
Our threat model assumes a black-box paraphraser without access to the watermark key, preserving watermark confidentiality. The adaptive attack \cite{diaa2025optimizing} we evaluated (Table~\ref{tab:adaptive_attack} in Appendix~\ref{app:adaptive_attack_app}) is optimized against the global z-score rather than PSS-specific signals (see also additional empirical evaluations in Appendix~\ref{app:failure_cases}); a fully PSS-adaptive attack would require white-box access to the feature design and joint optimization across windows and depths, a substantially stronger threat model that we leave to future work. Additionally, although we have expanded our analysis to three datasets (PG-19, CNN/DailyMail, WikiText), our empirical analysis is constrained to English long-form prose; low-entropy domains such as source code and highly technical writing are outside our current scope. Finally, supervised calibration may be sensitive to distribution shift, and threshold robustness across domains has yet to be systematically established.

\section*{Impact Statement}

This paper presents work whose goal is to advance the field of machine learning, specifically in the area of detecting machine-generated text through watermark detection. The primary societal benefit of this work is enabling more reliable identification of AI-generated content, which has applications in academic integrity, journalism verification, and content authenticity. While our detection methods could theoretically be studied by adversaries seeking to evade detection, we believe the benefits of improved detection capabilities outweigh these risks. The techniques we develop are detector-side improvements that work with existing watermarking infrastructure, requiring no changes to generation systems. We do not foresee specific negative societal consequences that must be highlighted beyond those that are well established when advancing detection capabilities for machine-generated text.

\bibliography{icml2026_watermark}

@inproceedings{kirchenbauer2023watermark,
  author={Kirchenbauer, John and Geiping, Jonas and Wen, Yuxin and Katz, Jonathan and Miers, Ian and Goldstein, Tom},
  title={A Watermark for Large Language Models},
  booktitle={International Conference on Machine Learning},
  pages={17061--17084},
  year={2023}
}

@inproceedings{fastdetectgpt,
  title={Fast-{D}etect{GPT}: Efficient Zero-Shot Detection of Machine-Generated Text via Conditional Probability Curvature},
  author={Bao, Guangsheng and Zhao, Yanbin and Teng, Zhiyang and Yang, Linyi and Zhang, Yue},
  booktitle={International Conference on Learning Representations},
  year={2024}
}

@InProceedings{detectgpt,
author =       {Mitchell, Eric and Lee, Yoonho and Khazatsky, Alexander and Manning, Christopher D and Finn, Chelsea},
title = 	 {{DetectGPT}: Zero-Shot Machine-Generated Text Detection using Probability Curvature},
booktitle = 	 {Proceedings of the 40th International Conference on Machine Learning},
pages = 	 {24950--24962},
year = 	 {2023},
volume = 	 {202},
series = 	 {Proceedings of Machine Learning Research}
}

@inproceedings{pg19,
  author={Rae, Jack W. and Potapenko, Anna and Jayakumar, Siddhant M. and Hillier, Chloe and Lillicrap, Timothy P.},
  title        = {Compressive Transformers for Long-Range Sequence Modelling},
  booktitle    = {International Conference on Learning Representations},
  year         = {2020}
}

@article{canreliablydetected,
author={Sadasivan, Vinu Sankar and Kumar, Aounon and Balasubramanian, Sriram and Wang, Wenxiao and Feizi, Soheil},
title={Can {AI}-Generated Text be Reliably Detected? {S}tress Testing {AI} Text Detectors Under Various Attacks},
journal={Transactions on Machine Learning Research},
issn={2835-8856},
year={2025}
}

@inproceedings{qu2025provably,
  author={Qu, Wenjie and Zheng, Wengrui and Tao, Tianyang and Yin, Dong and Jiang, Yanze and Tian, Zhihua and Zou, Wei and Jia, Jinyuan and Zhang, Jiaheng},
  title={Provably Robust Multi-bit Watermarking for {AI}-generated Text},
  booktitle={34th USENIX Security Symposium (USENIX Security 25)},
  pages={201--220},
  year={2025}
}

@inproceedings{he2025theoretically,
  author={He, Haiyun and Liu, Yepeng and Wang, Ziqiao and Mao, Yongyi and Bu, Yuheng},
  title={Theoretically Grounded Framework for {LLM} Watermarking: A Distribution-Adaptive Approach},
  booktitle={Advances in Neural Information Processing Systems},
  volume={38},
  year={2025}
}

@inproceedings{lau-etal-2024-waterfall,
author={Lau, Gregory Kang Ruey and Niu, Xinyuan and Dao, Hieu and Chen, Jiangwei and Foo, Chuan-Sheng and Low, Bryan Kian Hsiang},
title={Waterfall: Scalable Framework for Robust Text Watermarking and Provenance for {LLM}s},
booktitle={Proceedings of the 2024 Conference on Empirical Methods in Natural Language Processing},
pages={20432--20466},
year={2024},
doi={10.18653/v1/2024.emnlp-main.1138}
}

@article{feng2024certifiedrobustwatermarklarge,
  author={Feng, Xianheng and Liu, Jian and Ren, Kui and Chen, Chun},
  title={A Certified Robust Watermark for Large Language Models},
  journal={arXiv preprint arXiv:2409.19708},
  year={2024}
}

@inproceedings{linguistic_features,
  author={Gehrmann, Sebastian and Strobelt, Hendrik and Rush, Alexander},
  title={{GLTR}: Statistical Detection and Visualization of Generated Text},
  booktitle={Proceedings of the 57th Annual Meeting of the Association for Computational Linguistics: System Demonstrations},
  pages={111--116},
  year={2019},
  doi={10.18653/v1/P19-3019}
}

@article{style_embedding,
  author={Ding, Steven H. H. and Fung, Benjamin C. M. and Iqbal, Farkhund and Cheung, William K.},
  title={Learning Stylometric Representations for Authorship Analysis},
  journal={IEEE Transactions on Cybernetics},
  volume={49},
  number={1},
  pages={107--121},
  year={2019},
  doi={10.1109/TCYB.2017.2766189}
}

@inproceedings{biscope,
  author={Guo, Hanxi and Cheng, Siyuan and Jin, Xiaolong and Zhang, Zhuo and Zhang, Kaiyuan and Tao, Guanhong and Shen, Guangyu and Zhang, Xiangyu},
  title={{BiScope}: {AI}-generated Text Detection by Checking Memorization of Preceding Tokens},
  booktitle={Advances in Neural Information Processing Systems},
  volume={37},
  pages={104065--104090},
  year={2024}
}

@inproceedings{NEURIPS2023_575c4500,
 author={Krishna, Kalpesh and Song, Yixiao and Karpinska, Marzena and Wieting, John and Iyyer, Mohit},
  title={Paraphrasing evades detectors of {AI}-generated text, but retrieval is an effective defense},
  booktitle={Advances in Neural Information Processing Systems},
  volume={36},
  pages={27469--27500},
  year={2023}
}

@inproceedings{wang2024sourceattributionlargelanguage,
author={Lu, Xinyang and Wang, Jingtan and Zhao, Zitong and Dai, Zhongxiang and Foo, Chuan-Sheng and Ng, See-Kiong and Low, Bryan Kian Hsiang},
title={{WASA}: {WA}termark-based Source Attribution for Large Language Model-Generated Data},
booktitle={Findings of the Association for Computational Linguistics: ACL 2025},
pages={23791--23824},
year={2025},
doi={10.18653/v1/2025.findings-acl.1219}
}

@inproceedings{cheng2025adversarialparaphrasinguniversalattack,
title={Adversarial Paraphrasing: A Universal Attack for Humanizing {AI}-Generated Text},
author={Cheng, Yize and Sadasivan, Vinu Sankar and Saberi, Mehrdad and Saha, Shoumik and Feizi, Soheil},
booktitle={Advances in Neural Information Processing Systems},
year={2025}
}

@article{susnjak2022chatgpt,
  title={ChatGPT: The end of online exam integrity?},
  author={Susnjak, Teo and McIntosh, Timothy R},
  journal={Education Sciences},
  volume={14},
  number={6},
  pages={656},
  year={2024},
  publisher={MDPI}
}

@article{cotton2023chatting,
author={Cotton, Debby R. E. and Cotton, Peter A. and Shipway, J. Reuben},
title={Chatting and cheating: Ensuring academic integrity in the era of {ChatGPT}},
journal={Innovations in Education and Teaching International},
volume={61},
number={2},
pages={228--239},
year={2024},
doi={10.1080/14703297.2023.2190148}
}

@inproceedings{chen2024can,
  title={Can {LLM}-Generated Misinformation Be Detected?},
  author={Chen, Canyu and Shu, Kai},
  booktitle={International Conference on Learning Representations},
  year={2024}
}

@inproceedings{zhou2023synthetic,
author={Zhou, Jiawei and Zhang, Yixuan and Luo, Qianni and Parker, Andrea G. and De Choudhury, Munmun},
title={Synthetic Lies: Understanding {AI}-Generated Misinformation and Evaluating Algorithmic and Human Solutions},
booktitle={Proceedings of the 2023 CHI Conference on Human Factors in Computing Systems},
pages={1--20},
year = {2023},
doi = {10.1145/3544548.3581318}
}

@article{nas2024protecting,
author = {Wolfgang Blau  and Vinton G. Cerf  and Juan Enriquez  and Joseph S. Francisco  and Urs Gasser  and Mary L. Gray  and Mark Greaves  and Barbara J. Grosz  and Kathleen Hall Jamieson  and Gerald H. Haug  and John L. Hennessy  and Eric Horvitz  and David I. Kaiser  and Alex John London  and Robin Lovell-Badge  and Marcia K. McNutt  and Martha Minow  and Tom M. Mitchell  and Susan Ness  and Shobita Parthasarathy  and Saul Perlmutter  and William H. Press  and Jeannette M. Wing  and Michael Witherell },
title = {Protecting scientific integrity in an age of generative {AI}},
journal = {Proceedings of the National Academy of Sciences},
volume = {121},
number = {22},
pages = {e2407886121},
year = {2024},
doi = {10.1073/pnas.2407886121}
}

@article{gao2023comparing,
  author={Gao, Catherine A. and Howard, Frederick M. and Markov, Nikolay S. and Dyer, Emma C. and Ramesh, Siddhi and Luo, Yuan and Pearson, Alexander T.},
  title={Comparing scientific abstracts generated by {ChatGPT} to real abstracts with detectors and blinded human reviewers},
  journal={npj Digital Medicine},
  volume={6},
  number={1},
  pages={75},
  year={2023},
  doi={10.1038/s41746-023-00819-6}
}

@article{dathathri2024scalable,
  author={Dathathri, Sumanth and See, Abigail and Ghaisas, Sumedh and Huang, Po-Sen and McAdam, Rob and Welbl, Johannes and Bachani, Vandana and Kaskasoli, Alex and Stanforth, Robert and Matejovicova, Tatiana and Hayes, Jamie and Vyas, Nidhi and Merey, Majd Al and Brown-Cohen, Jonah and Bunel, Rudy and Balle, Borja and Cemgil, Taylan and Ahmed, Zahra and Stacpoole, Kitty and Shumailov, Ilia and Baetu, Ciprian and Gowal, Sven and Hassabis, Demis and Kohli, Pushmeet},
  title={Scalable watermarking for identifying large language model outputs},
  journal={Nature},
  volume={634},
  number={8035},
  pages={818--823},
  year={2024},
  doi={10.1038/s41586-024-08025-4}
}

@inproceedings{rastogi2024revisiting,
  author={Rastogi, Saksham and Pruthi, Danish},
  title        = {Revisiting the Robustness of Watermarking to Paraphrasing Attacks},
  booktitle    = {Proceedings of the 2024 Conference on Empirical Methods in Natural Language Processing},
  pages={18100--18110},
  year         = {2024},
  doi={10.18653/v1/2024.emnlp-main.1005}
}

@inproceedings{semamark2023,
  author={Ren, Jie and Xu, Han and Liu, Yiding and Cui, Yingqian and Wang, Shuaiqiang and Yin, Dawei and Tang, Jiliang},
  title={A Robust Semantics-based Watermark for Large Language Model against Paraphrasing},
  booktitle={Findings of the Association for Computational Linguistics: NAACL 2024},
  pages={613--625},
  year={2024},
  doi={10.18653/v1/2024.findings-naacl.40}
}

@article{fairoze2025publicly,
  author={Fairoze, Jaiden and Garg, Sanjam and Jha, Somesh and Mahloujifar, Saeed and Mahmoody, Mohammad and Wang, Mingyuan},
  title={Publicly-Detectable Watermarking for Language Models},
  journal={IACR Communications in Cryptology},
  volume={1},
  number={4},
  year={2025},
  doi={10.62056/ahmpdkp10}
}

@inproceedings{saemark2024,
  title={SAEMark: Steering Personalized Multilingual LLM Watermarks with Sparse Autoencoders},
  author={Yu, Zhuohao and Jiang, Xingru and Gu, Weizheng and Wang, Yidong and Wen, Qingsong and Zhang, Shikun and Ye, Wei},
  booktitle={Advances in Neural Information Processing Systems (NeurIPS)},
  year={2025}
}

@article{xu2024freqmarkfrequencybasedwatermarksentencelevel,
author={Xu, Zhenyu and Zhang, Kun and Sheng, Victor S.},
title={{FreqMark}: Frequency-Based Watermark for Sentence-Level Detection of {LLM}-Generated Text},
journal={arXiv preprint arXiv:2410.10876},
year={2024}
}

@inproceedings{adaptivetext2024,
  author={Liu, Yepeng and Bu, Yuheng},
  title={Adaptive Text Watermark for Large Language Models},
  booktitle={International Conference on Machine Learning},
  pages={30718--30737},
  year={2024}
}

@inproceedings{ensemble2024,
  author={Niess, Georg and Kern, Roman},
  title={Ensemble Watermarks for Large Language Models},
  booktitle={Proceedings of the 63rd Annual Meeting of the Association for Computational Linguistics (Volume 1: Long Papers)},
  pages={2903--2916},
  year={2025},
  doi={10.18653/v1/2025.acl-long.145}
}

@inproceedings{li2025likelihood,
  author = {Li, Xingchi and Li, Guanxun and Zhang, Xianyang},
  title = {A Likelihood Based Approach for Watermark Detection},
  booktitle = {Proceedings of The 28th International Conference on Artificial Intelligence and Statistics},
  pages = {1675--1683},
  year = {2025},
  volume = {258},
  series = {Proceedings of Machine Learning Research}
}

@inproceedings{diaa2025optimizing,
author={Diaa, Abdulrahman and Aremu, Toluwani and Lukas, Nils},
title={Optimizing Adaptive Attacks against Watermarks for Language Models},
booktitle={Proceedings of the 42nd International Conference on Machine Learning},
year={2025},
volume={267},
series={Proceedings of Machine Learning Research},
publisher={PMLR}
}

@inproceedings{he2024watermarks,
  author={He, Zhiwei and Zhou, Binglin and Hao, Hongkun and Liu, Aiwei and Wang, Xing and Tu, Zhaopeng and Zhang, Zhuosheng and Wang, Rui},
  title={Can Watermarks Survive Translation? On the Cross-lingual Consistency of Text Watermark for Large Language Models},
  booktitle={Proceedings of the 62nd Annual Meeting of the Association for Computational Linguistics (Volume 1: Long Papers)},
  pages={4115--4129},
  year={2024},
  doi={10.18653/v1/2024.acl-long.226}
}

@article{evaluating2024,
author={Liu, Zesen and Cong, Tianshuo and He, Xinlei and Li, Qi},
title={On Evaluating The Performance of Watermarked Machine-Generated Texts Under Adversarial Attacks}, 
year={2024},
journal={arXiv preprint arXiv:2407.04794}
}

@inproceedings{lee-etal-2024-wrote,
  author={Lee, Taehyun and Hong, Seokhee and Ahn, Jaewoo and Hong, Ilgee and Lee, Hwaran and Yun, Sangdoo and Shin, Jamin and Kim, Gunhee},
  title={Who Wrote this Code? Watermarking for Code Generation},
  booktitle={Proceedings of the 62nd Annual Meeting of the Association for Computational Linguistics (Volume 1: Long Papers)},
  pages={4890--4911},
  year={2024},
  doi={10.18653/v1/2024.acl-long.268}
}

@inproceedings{hastuti2025factuality,
  author={Hastuti, Rochana Prih and Rajagede, Rian Adam and Al Ghanim, Mansour and Zheng, Mengxin and Lou, Qian},
  title={Factuality Beyond Coherence: Evaluating {LLM} Watermarking Methods for Medical Texts},
  booktitle={Findings of the Association for Computational Linguistics: EMNLP 2025},
  pages={15129--15147},
  year={2025}
}

@article{statframework2024,
author = {Xiang Li and Feng Ruan and Huiyuan Wang and Qi Long and Weijie J. Su},
title = {{A statistical framework of watermarks for large language models: Pivot, detection efficiency and optimal rules}},
volume = {53},
journal = {The Annals of Statistics},
number = {1},
pages = {322 -- 351},
year = {2025},
doi = {10.1214/24-AOS2468}
}

@article{majormark2024,
author={Xu, Jiahao and Hu, Rui and Zhang, Zikai},
title={Majority Bit-Aware Watermarking for Large Language Models},
journal={arXiv preprint arXiv:2508.03829},
year={2025}
}

@inproceedings{semanticinvariant2024,
  author={Liu, Aiwei and Pan, Leyi and Hu, Xuming and Meng, Shiao and Wen, Lijie},
  title={A Semantic Invariant Robust Watermark for Large Language Models},
  booktitle={The Twelfth International Conference on Learning Representations},
  year={2024}
}

@inproceedings{feng2025bimark,
  author={Feng, Xiaoyan and Zhang, He and Zhang, Yanjun and Zhang, Leo Yu and Pan, Shirui},
  title={{BiMark}: Unbiased Multilayer Watermarking for Large Language Models},
  booktitle={International Conference on Machine Learning},
  year={2025}
}

@inproceedings{zellers2019grover,
author={Zellers, Rowan and Holtzman, Ari and Rashkin, Hannah and Bisk, Yonatan and Farhadi, Ali and Roesner, Franziska and Choi, Yejin},
title={Defending Against Neural Fake News},
booktitle={Advances in Neural Information Processing Systems},
volume={32},
pages={9054--9065},
year={2019}
}

@inproceedings{hans2024binoculars,
  author={Hans, Abhimanyu and Schwarzschild, Avi and Cherepanova, Valeriia and Kazemi, Hamid and Saha, Aniruddha and Goldblum, Micah and Geiping, Jonas and Goldstein, Tom},
  title={Spotting {LLMs} With Binoculars: Zero-Shot Detection of Machine-Generated Text},
  booktitle={International Conference on Machine Learning},
  pages={17519--17537},
  year={2024}
}

@article{deeptextmark2024,
  author={Munyer, Travis and Tanvir, Abdullah All and Das, Arjon and Zhong, Xin},
  title={{DeepTextMark}: A Deep Learning-Driven Text Watermarking Approach for Identifying Large Language Model Generated Text}, 
  journal={IEEE Access}, 
  year={2024},
  volume={12},
  pages={40508-40520},
  doi={10.1109/ACCESS.2024.3376693}
}

@inproceedings{radar2024,
 author={Hu, Xiaomeng and Chen, Pin-Yu and Ho, Tsung-Yi},
  title={{RADAR}: Robust {AI}-Text Detection via Adversarial Learning},
  booktitle={Advances in Neural Information Processing Systems},
  volume={36},
  pages={15077--15095},
  year={2023}
}

@inproceedings{wikitext,
  author={Merity, Stephen and Xiong, Caiming and Bradbury, James and Socher, Richard},
  title={Pointer Sentinel Mixture Models},
  booktitle={International Conference on Learning Representations},
  year={2017}
}

@inproceedings{NIPS2015_afdec700,
 author={Hermann, Karl Moritz and Kocisky, Tomas and Grefenstette, Edward and Espeholt, Lasse and Kay, Will and Suleyman, Mustafa and Blunsom, Phil},
  title={Teaching Machines to Read and Comprehend},
  booktitle={Advances in Neural Information Processing Systems},
  volume={28},
  pages={1693--1701},
  year={2015}
}

@inproceedings{kirchenbauer2024reliability,
  author={Kirchenbauer, John and Geiping, Jonas and Wen, Yuxin and Shu, Manli and Saifullah, Khalid and Kong, Kezhi and Fernando, Kasun and Saha, Aniruddha and Goldblum, Micah and Goldstein, Tom},
  title={On the Reliability of Watermarks for Large Language Models},
  booktitle={International Conference on Learning Representations},
  year={2024}
}

@inproceedings{cheng2025sira,
  author={Cheng, Yixin and Guo, Hongcheng and Li, Yangming and Sigal, Leonid},
  title={Revealing Weaknesses in Text Watermarking Through Self-Information Rewrite Attacks},
  booktitle={International Conference on Machine Learning},
  year={2025}
}

@article{kuditipudi2024robust,
  author={Kuditipudi, Rohith and Thickstun, John and Hashimoto, Tatsunori and Liang, Percy},
  title={Robust Distortion-free Watermarks for Language Models},
  journal={Transactions on Machine Learning Research},
  issn={2835-8856},
  year={2024}
}

@misc{aaronson2022watermarking,
  author={Aaronson, Scott},
  title={My {AI} Safety Lecture for {UT} Effective Altruism},
  year={2022},
  howpublished={Blog post},
  note={\url{https://scottaaronson.blog/?p=6823}}
}

@inproceedings{christ2024undetectable,
  author={Christ, Miranda and Gunn, Sam and Zamir, Or},
  title={Undetectable Watermarks for Language Models},
  booktitle={Proceedings of Thirty Seventh Conference on Learning Theory},
  pages={1125--1139},
  year={2024},
  volume={247},
  series={Proceedings of Machine Learning Research},
  publisher={PMLR}
}

@inproceedings{jovanovic2024watermarkstealing,
  author={Jovanovi{\'c}, Nikola and Staab, Robin and Vechev, Martin},
  title={Watermark Stealing in Large Language Models},
  booktitle={International Conference on Machine Learning},
  year={2024}
}

@inproceedings{zhao2025efficiently,
  author={Zhao, Xuandong and Liao, Chenwen and Wang, Yu-Xiang and Li, Lei},
  title={Efficiently Identifying Watermarked Segments in Mixed-Source Texts},
  booktitle={Proceedings of the 63rd Annual Meeting of the Association for Computational Linguistics (Volume 1: Long Papers)},
  pages={6304--6316},
  year={2025},
  doi={10.18653/v1/2025.acl-long.316}
}

@article{xie2025detecting,
  author={Xie, Liyan and Siddeek, Muhammad and Seif, Mohamed and Goldsmith, Andrea J. and Wang, Mengdi},
  title={Detecting Post-generation Edits to Watermarked {LLM} Outputs via Combinatorial Watermarking},
  journal={arXiv preprint arXiv:2510.01637},
  year={2025}
}

@inproceedings{shen2025enhancing,
  author={Shen, Huanming and Huang, Baizhou and Wan, Xiaojun},
  title={Enhancing {LLM} Watermark Resilience Against Both Scrubbing and Spoofing Attacks},
  booktitle={Advances in Neural Information Processing Systems},
  year={2025}
}

@article{emi2024pangram,
  author={Emi, Bradley and Spero, Max},
  title={Technical Report on the {Pangram} {AI}-Generated Text Classifier},
  journal={arXiv preprint arXiv:2402.14873},
  year={2024}
}
\bibliographystyle{icml2026}

%%%%%%%%%%%%%%%%%%%%%%%%%%%%%%%%%%%%%%%%%%%%%%%%%%%%%%%%%%%%%%%%%%%%%%%%%%%%%%%
%%%%%%%%%%%%%%%%%%%%%%%%%%%%%%%%%%%%%%%%%%%%%%%%%%%%%%%%%%%%%%%%%%%%%%%%%%%%%%%
% APPENDIX
%%%%%%%%%%%%%%%%%%%%%%%%%%%%%%%%%%%%%%%%%%%%%%%%%%%%%%%%%%%%%%%%%%%%%%%%%%%%%%%
%%%%%%%%%%%%%%%%%%%%%%%%%%%%%%%%%%%%%%%%%%%%%%%%%%%%%%%%%%%%%%%%%%%%%%%%%%%%%%%
\newpage
\appendix
\onecolumn
\section{Appendix}
\label{sec:appendix}

In this appendix, we provide complementary material to support the main text. 
Section~\ref{app:paraphraser_sensitivity} reports initial sensitivity tests, including robustness to different paraphrasers (Gemma-7B-IT and Qwen2-7B-Instruct) while additional sensitivity results including window/stride, classifier choice, watermark schemes, shorter texts, and mixed paraphrasing are provided in Section~\ref{app:more_sensitivity}. Section~\ref{app:practical_metrics} provides other evaluation metrics, semantic preservation results, and additional attack scenarios.
% \textcolor{blue}{Section~\ref{app:cross_domain_detailed} provides detailed cross-domain and within-domain results across all datasets.}
Section~\ref{app:compute} details compute and implementation notes, and it also includes dataset-level pseudocode matching our implementation. Section~\ref{app:failure_cases} documents failure cases and limits of detectability, and Section~\ref{app:adaptive_attack_app} presents the full adaptive attack evaluation.

\paragraph{Setup recap.}
We evaluate four families of detectors: (1) \emph{Global z threshold}; (2) \emph{Local z (20)} (first 20 rolling-window local z's as features); (3) \emph{Static features} (windowed moments, short-range autocorrelations, run-length and run-frequency summaries); and (4) \emph{PSS + static} which augments static features with \emph{Pattern Stability Scores} computed from standard deviation of aligned local z-score trajectories across depths. All models use XGBoost for classifier-based lines unless stated; windows use $w{=}50$, stride $s{=}10$, with the non-fragmenting expansion rule. This mirrors the main-text configuration to avoid confounds. Paraphrases are generated up to $D9$ to  enable PSS computation at the maximum evaluation depth $D8$.

\subsection{Sensitivity to Paraphraser Choice}
\label{app:paraphraser_sensitivity}

\paragraph{Motivation.}
Retroactive detectors sometimes latch onto paraphraser-specific artifacts. Our detector explicitly targets \emph{cross-depth stability} of local watermark evidence, which should persist irrespective of the paraphrasing model. We therefore re-run the entire pipeline with two single-model paraphrasers beyond Mistral-7B-Instruct used in the main text: \emph{Gemma-7B-IT} and \emph{Qwen2-7B-Instruct}. For each, we generate $D1$--$D8$ chains under the same prompts and decoding settings as in Section~\ref{sec:experiments}.

\paragraph{Results.}
Across both paraphrasers, the ordering of methods is consistent with the main text: global thresholding drops fastest with depth; injecting locality slows degradation; and \emph{PSS + static} yields the flattest curves and the highest accuracies at mid/late depths. In particular, the stability signal is additive to locality, preserving margins even when token distributions shift due to a different rewriting policy. Full AUC values per depth are reported in Tables~\ref{tab:gemma_auc_meanstd} and \ref{tab:qwen2_auc_meanstd}.

\begin{table}[h]
\centering
\caption{\textbf{Gemma-7B-IT paraphrasing: AUC (\%) vs.\ depth ($D1$--$D8$).} All classifier entries use XGBoost; values are mean $\pm$ std over 30 runs.}
\vspace{2pt}
\resizebox{\textwidth}{!}{%
\begin{tabular}{lcccccccc}
\toprule
\textbf{Method} & \textbf{D1} & \textbf{D2} & \textbf{D3} & \textbf{D4} & \textbf{D5} & \textbf{D6} & \textbf{D7} & \textbf{D8} \\
\midrule
Global z-score threshold & 56.50 $\pm$ 0.00 & 54.35 $\pm$ 0.00 & 53.75 $\pm$ 0.00 & 53.55 $\pm$ 0.00 & 53.25 $\pm$ 0.00 & 53.30 $\pm$ 0.00 & 53.05 $\pm$ 0.00 & 53.00 $\pm$ 0.00 \\
Local z-score (20) & 66.07 $\pm$ 1.42 & 59.03 $\pm$ 1.66 & 58.92 $\pm$ 1.99 & 57.52 $\pm$ 2.20 & 57.81 $\pm$ 1.63 & 57.32 $\pm$ 1.40 & 57.55 $\pm$ 1.55 & 57.80 $\pm$ 1.41 \\
Static features & 72.75 $\pm$ 1.87 & 68.95 $\pm$ 1.87 & 67.02 $\pm$ 1.29 & 68.10 $\pm$ 1.35 & 67.48 $\pm$ 1.76 & 67.10 $\pm$ 1.83 & 66.82 $\pm$ 1.58 & 65.16 $\pm$ 1.87 \\
\textbf{PSS + Static} & \textbf{83.51} $\pm$ \textbf{1.43} & \textbf{77.85} $\pm$ \textbf{1.63} & \textbf{75.98} $\pm$ \textbf{1.45} & \textbf{76.54} $\pm$ \textbf{1.21} & \textbf{73.69} $\pm$ \textbf{1.49} & \textbf{73.33} $\pm$ \textbf{1.23} & \textbf{72.92} $\pm$ \textbf{2.03} & -- \\
\bottomrule
\end{tabular}%
}
\label{tab:gemma_auc_meanstd}
\end{table}

\begin{table}[h]
\centering
\caption{\textbf{Qwen2-7B-Instruct paraphrasing: AUC (\%) vs.\ depth ($D1$--$D8$).} All classifier entries use XGBoost; values are mean $\pm$ std over 30 runs.}
\vspace{2pt}
\resizebox{\textwidth}{!}{%
\begin{tabular}{lcccccccc}
\toprule
\textbf{Method} & \textbf{D1} & \textbf{D2} & \textbf{D3} & \textbf{D4} & \textbf{D5} & \textbf{D6} & \textbf{D7} & \textbf{D8} \\
\midrule
Global z-score threshold & 73.25 $\pm$ 0.00 & 69.90 $\pm$ 0.00 & 69.00 $\pm$ 0.00 & 68.40 $\pm$ 0.00 & 67.90 $\pm$ 0.00 & 67.70 $\pm$ 0.00 & 67.35 $\pm$ 0.00 & 66.80 $\pm$ 0.00 \\
Local z-score (20) & 58.90 $\pm$ 1.61 & 58.15 $\pm$ 1.69 & 55.03 $\pm$ 1.71 & 55.48 $\pm$ 1.64 & 55.86 $\pm$ 1.66 & 55.19 $\pm$ 1.64 & 54.27 $\pm$ 1.63 & 55.27 $\pm$ 1.66 \\
Static features & 82.08 $\pm$ 1.45 & 78.64 $\pm$ 1.55 & 77.15 $\pm$ 1.51 & 77.89 $\pm$ 1.48 & 77.47 $\pm$ 1.52 & 78.37 $\pm$ 1.50 & 78.77 $\pm$ 1.49 & 76.36 $\pm$ 1.47 \\
\textbf{PSS + Static} & \textbf{94.67} $\pm$ \textbf{0.68} & \textbf{92.68} $\pm$ \textbf{0.79} & \textbf{92.38} $\pm$ \textbf{0.82} & \textbf{92.63} $\pm$ \textbf{0.78} & \textbf{91.90} $\pm$ \textbf{0.91} & \textbf{90.57} $\pm$ \textbf{0.98} & \textbf{90.47} $\pm$ \textbf{1.02} & -- \\
\bottomrule
\end{tabular}%
}
\label{tab:qwen2_auc_meanstd}
\end{table}

\paragraph{Takeaway.}
Consistent rankings across paraphrasers support the claim that stability-aware local detection is \emph{paraphraser-agnostic}. The detector exploits invariants (local concentration and cross-depth persistence) that are difficult to erase simultaneously without semantic drift or length distortion.

\subsection{Additional Sensitivities}
\label{app:more_sensitivity}

Tables~\ref{tab:within_cnn} and~\ref{tab:within_wiki} report within-domain results when training and testing on CNN/DailyMail and WikiText respectively, while Tables~\ref{tab:cross_wiki_cnn} and~\ref{tab:cross_cnn_wiki} report the remaining cross-domain pairings (WikiText$\to$CNN/DailyMail and CNN/DailyMail$\to$WikiText) that complement Table~\ref{tab:cross_domain} in the main text. Table~\ref{tab:comprehensive} consolidates PSS\,+\,Static performance across all dataset/LLM/paraphraser/$\gamma$ combinations evaluated in this paper. In all cases, the same qualitative ordering observed in the main text holds: \emph{PSS\,+\,Static} dominates, followed by static features, then local-z and global-z baselines.

\begin{figure}[t]
\centering
\includegraphics[width=0.44\linewidth]{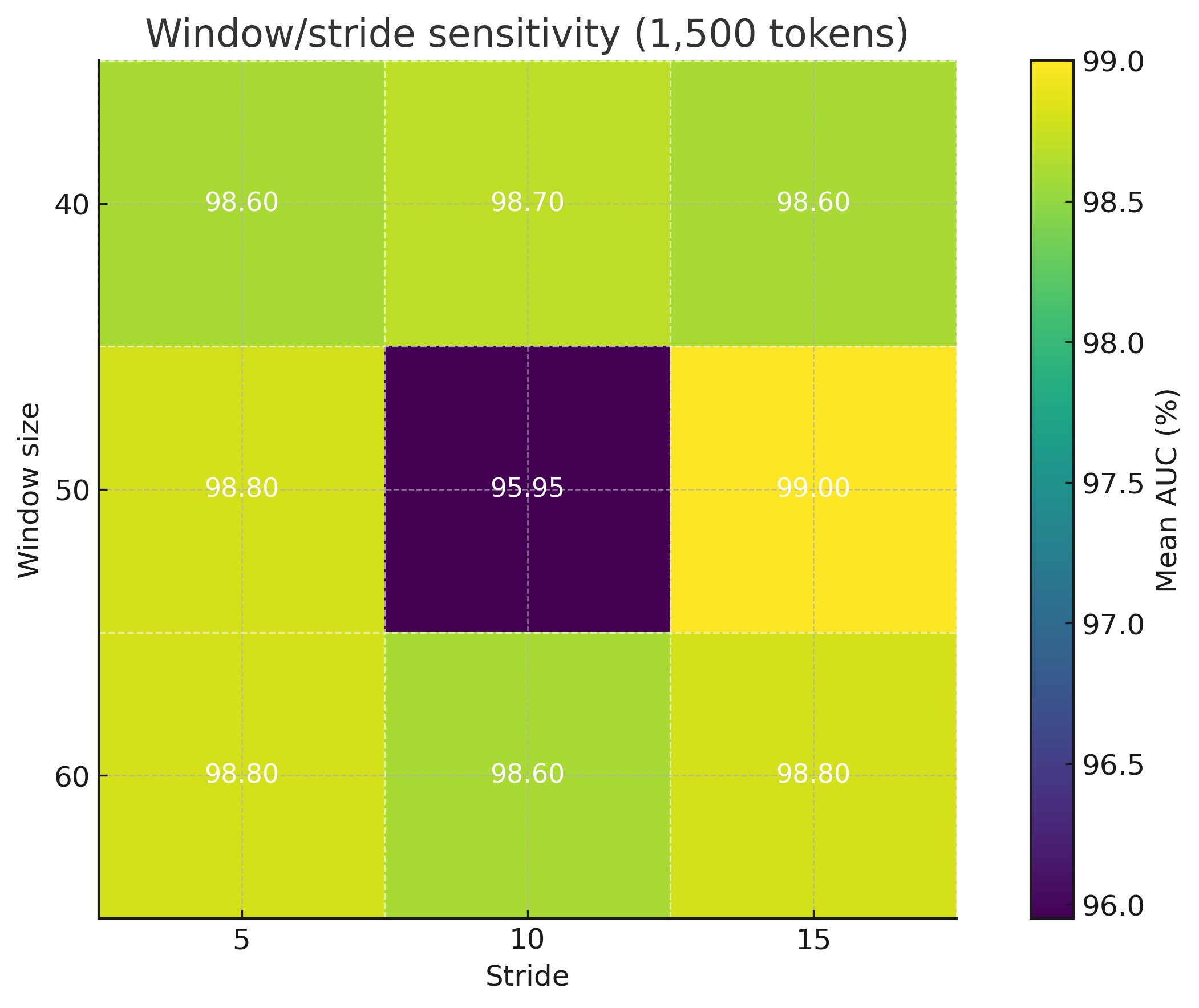}
\caption{\textbf{Window/stride sensitivity (1{,}500 tokens)} for \emph{PSS + static}. Numbers show mean AUC (\%) across depths $D1$--$D8$.}
\label{fig:heatmap_ws_1500_auc}
\end{figure}

\paragraph{Window and stride.} Figure~\ref{fig:heatmap_ws_1500_auc} summarizes the \emph{mean} AUC across depths ($D1$--$D8$) for all $(w,s)\in{ \lbrace40,50,60 \rbrace}\times{ \lbrace5,10,15 \rbrace}$ at 1{,}500 tokens. Performance is highly stable: the best setting $(50,15)$ achieves 99.0\%, while the lowest $(50,10)$ records 95.95\%, a spread of only 3.05 percentage points. The configuration $(50,10)$ used throughout yields 95.95\%, lying well within this plateau, confirming that our detector remains robust to moderate changes in window size and stride.

\paragraph{Classifier choice.}
Using LR/RF/XGB/SVM/$k$NN for local/static features yields the same ordering. XGBoost is typically best. Gains primarily trace to feature design rather than model complexity.

\paragraph{Feature-group ablation.} Table~\ref{tab:feature_ablation} reports the incremental contribution of each feature group, starting from the global z-score baseline and adding components one at a time, evaluated at depths $D1$, $D3$, $D5$, and $D8$ on PG-19 at 1{,}500 tokens. This complements the textual summary in the main paper by showing the per-depth numerical breakdown.

\begin{table}[h]
\centering
\caption{\textbf{Feature-group ablation.} AUC (\%) on PG-19 at 1{,}500 tokens. Components are added incrementally to a global z-score baseline. Every group contributes measurably and PSS provides the largest single jump, confirming that each component captures a distinct, non-redundant aspect of the watermark signal.}
\vspace{2pt}
\resizebox{0.65\textwidth}{!}{%
\begin{tabular}{lcccc}
\toprule
\textbf{Feature Set} & \textbf{D1} & \textbf{D3} & \textbf{D5} & \textbf{D8} \\
\midrule
Global z-score (baseline) & 74.2 & 70.0 & 68.0 & 66.8 \\
Z-score moments (6-D) & 83.2 & 76.5 & 74.1 & 72.0 \\
~~+ Autocorrelations (8-D) & 85.0 & 79.3 & 76.0 & 74.4 \\
~~+ Run-length stats (14-D) & 87.3 & 83.2 & 79.8 & 77.2 \\
~~+ Run-frequency (full 20-D) & 88.1 & 83.2 & 80.2 & 78.6 \\
\textbf{~~+ PSS (full PSS\,+\,Static)} & \textbf{96.1} & \textbf{93.9} & \textbf{92.6} & \textbf{91.2} \\
\bottomrule
\end{tabular}%
}
\label{tab:feature_ablation}
\end{table}

\paragraph{Watermarking scheme compatibility.}
% To demonstrate that PSS is watermark-agnostic, we evaluate with different greenlist ratios (Table~\ref{tab:watermark_schemes}). With standard $\gamma$=0.25, we achieve 92.1\% at D1 and 86.3\% at D8. With stronger $\gamma$=0.5, performance improves to 94.8\% at D1 and 89.9\% at D8. This confirms PSS captures fundamental watermark properties rather than scheme-specific artifacts, enabling deployment with various existing watermarking configurations.
To demonstrate that PSS is watermark-agnostic, we evaluate with different greenlist ratios (Table~\ref{tab:watermark_schemes}). With standard $\gamma$=0.25, we achieve 96.1\% AUC at D1 and 91.2\% at D8. With stronger $\gamma$=0.5, performance improves to 97.6\% at D1 and 94.2\% at D8. This confirms PSS captures fundamental watermark properties rather than scheme-specific artifacts, enabling deployment with various existing watermarking configurations.

\begin{table}[h]
\centering
\caption{\textbf{PSS with Different Watermarking Schemes.} AUC (\%) with varying greenlist ratios.}
\vspace{2pt}
\resizebox{0.5\textwidth}{!}{%
\begin{tabular}{lcccc}
\toprule
\textbf{Watermark Config} & \textbf{D1} & \textbf{D3} & \textbf{D5} & \textbf{D8} \\
\midrule
Standard ($\gamma$=0.25) & 96.1 & 93.9 & 92.6 & 91.2 \\
Stronger ($\gamma$=0.50) & 97.6 & 96.4 & 95.5 & 94.2 \\
\bottomrule
\end{tabular}%
}
\label{tab:watermark_schemes}
\end{table}

% \begin{table}[h]
% \centering
% \caption{\textbf{PSS with Different Watermarking Schemes.} Performance (\%) with varying greenlist ratios.}
% \vspace{2pt}
% \resizebox{0.5\textwidth}{!}{%
% \begin{tabular}{lcccc}
% \toprule
% \textbf{Watermark Config} & \textbf{D1} & \textbf{D3} & \textbf{D5} & \textbf{D8} \\
% \midrule
% Standard ($\gamma$=0.25) & 92.1 & 89.2 & 87.8 & 86.3 \\
% Stronger ($\gamma$=0.50) & 94.8 & 92.7 & 91.4 & 89.9 \\
% \bottomrule
% \end{tabular}%
% }
% \label{tab:watermark_schemes}
% \end{table}

\begin{figure}[t]
\vskip 0.2in
\begin{center}
\begin{minipage}{0.49\columnwidth}
\centering
\includegraphics[width=\linewidth]{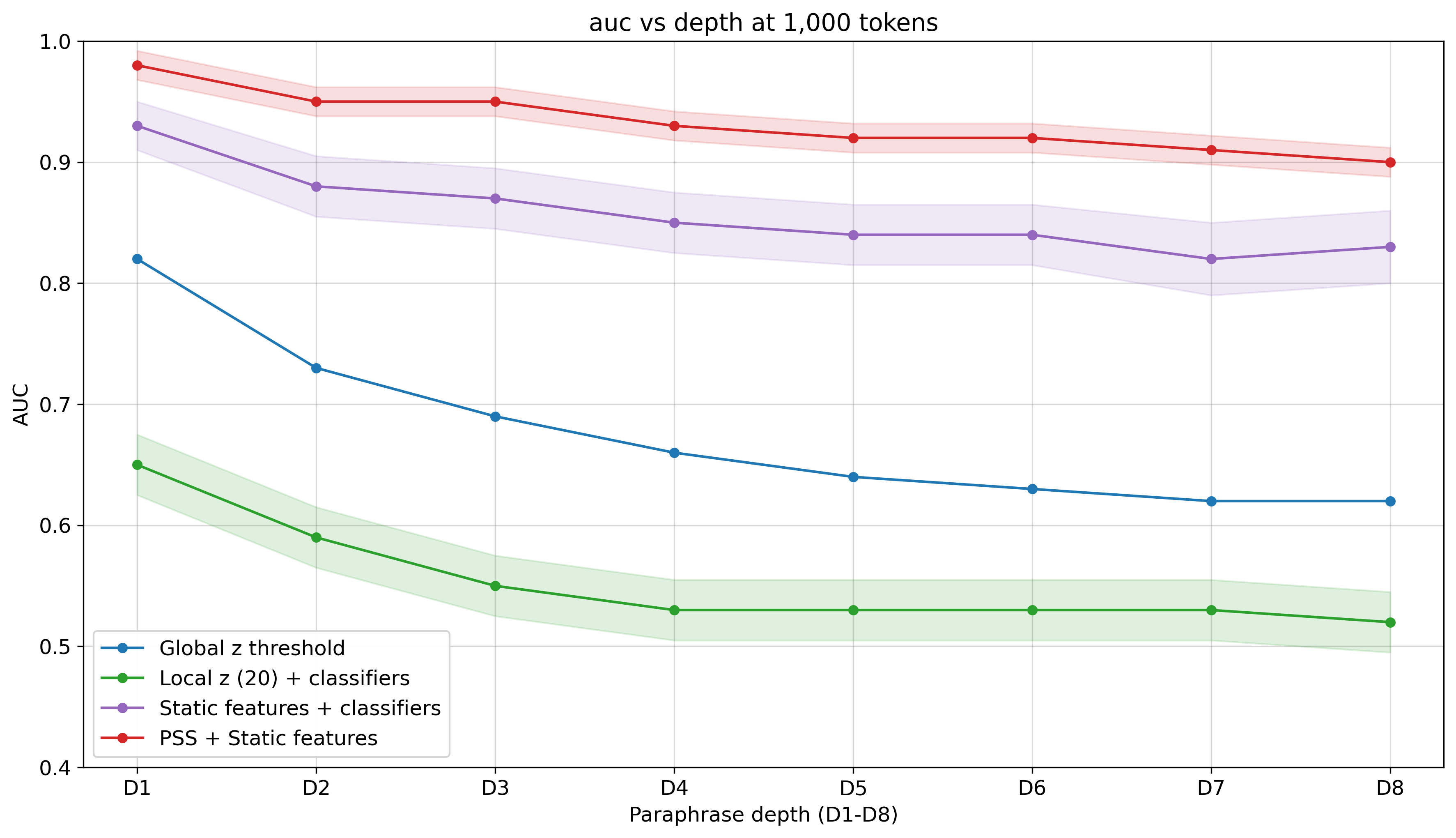}\\[2pt]
\includegraphics[width=\linewidth]{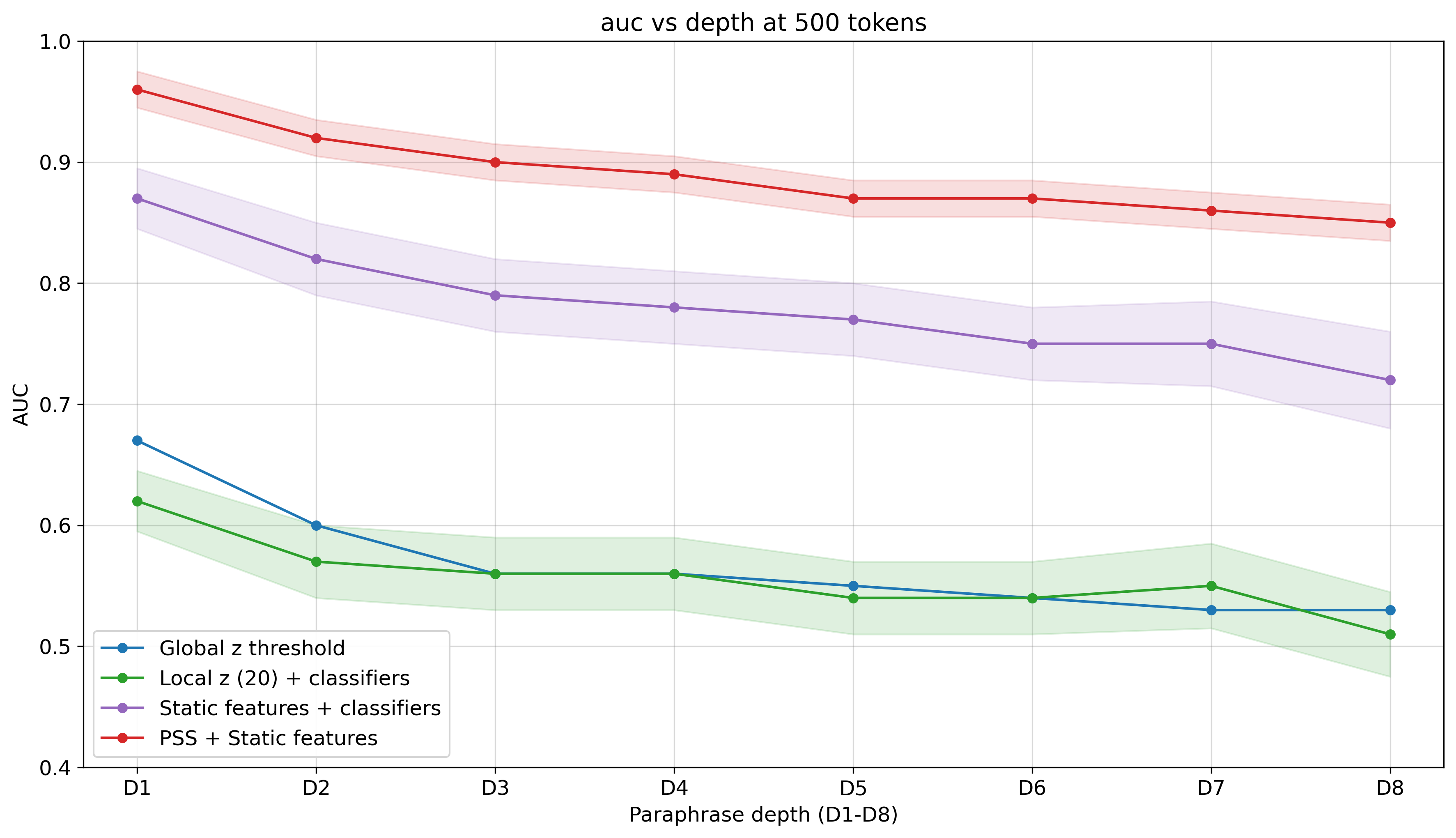}
\end{minipage}\hfill
\begin{minipage}{0.49\columnwidth}
\centering
\includegraphics[width=\linewidth]{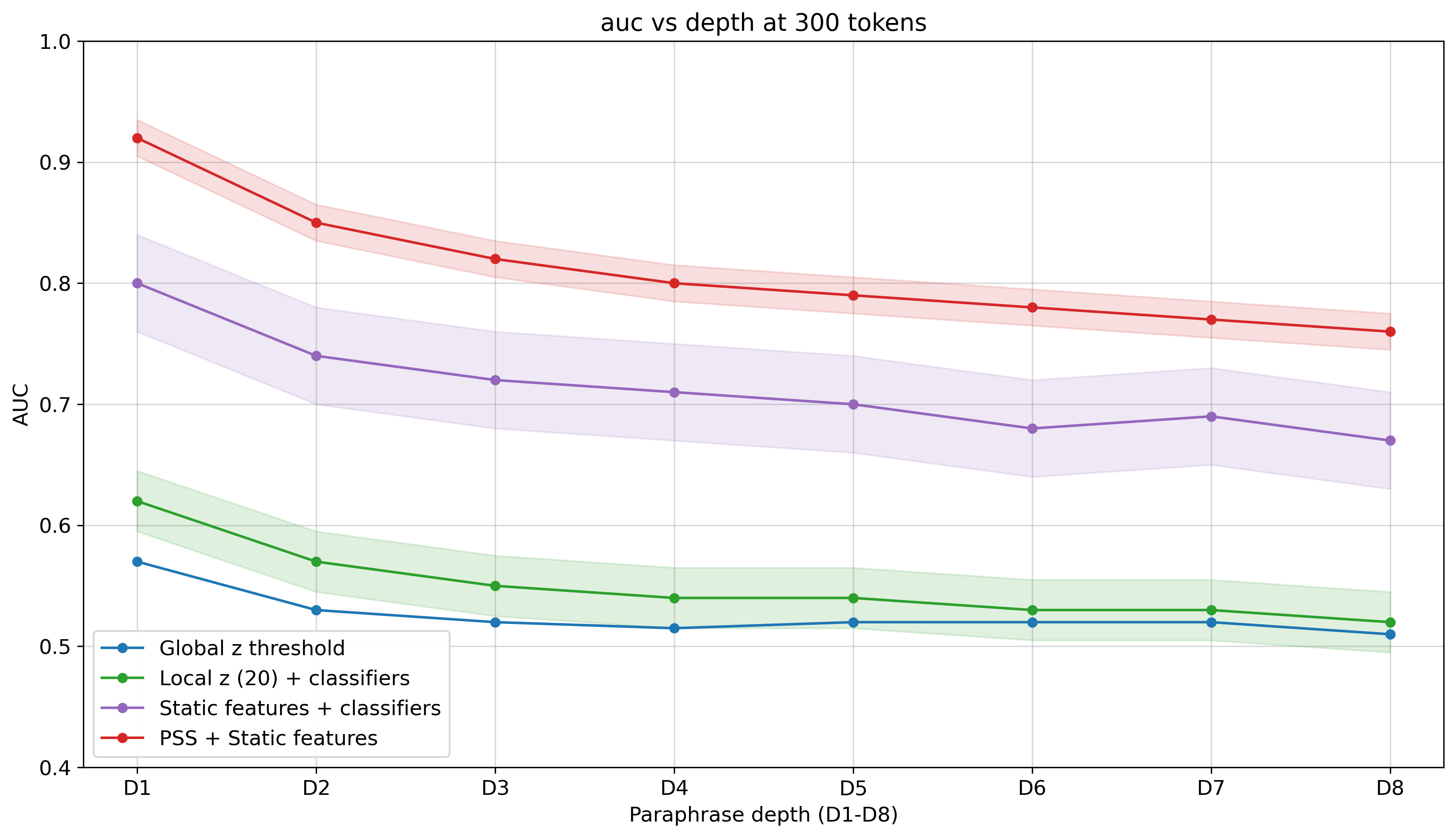}\\[2pt]
\includegraphics[width=\linewidth]{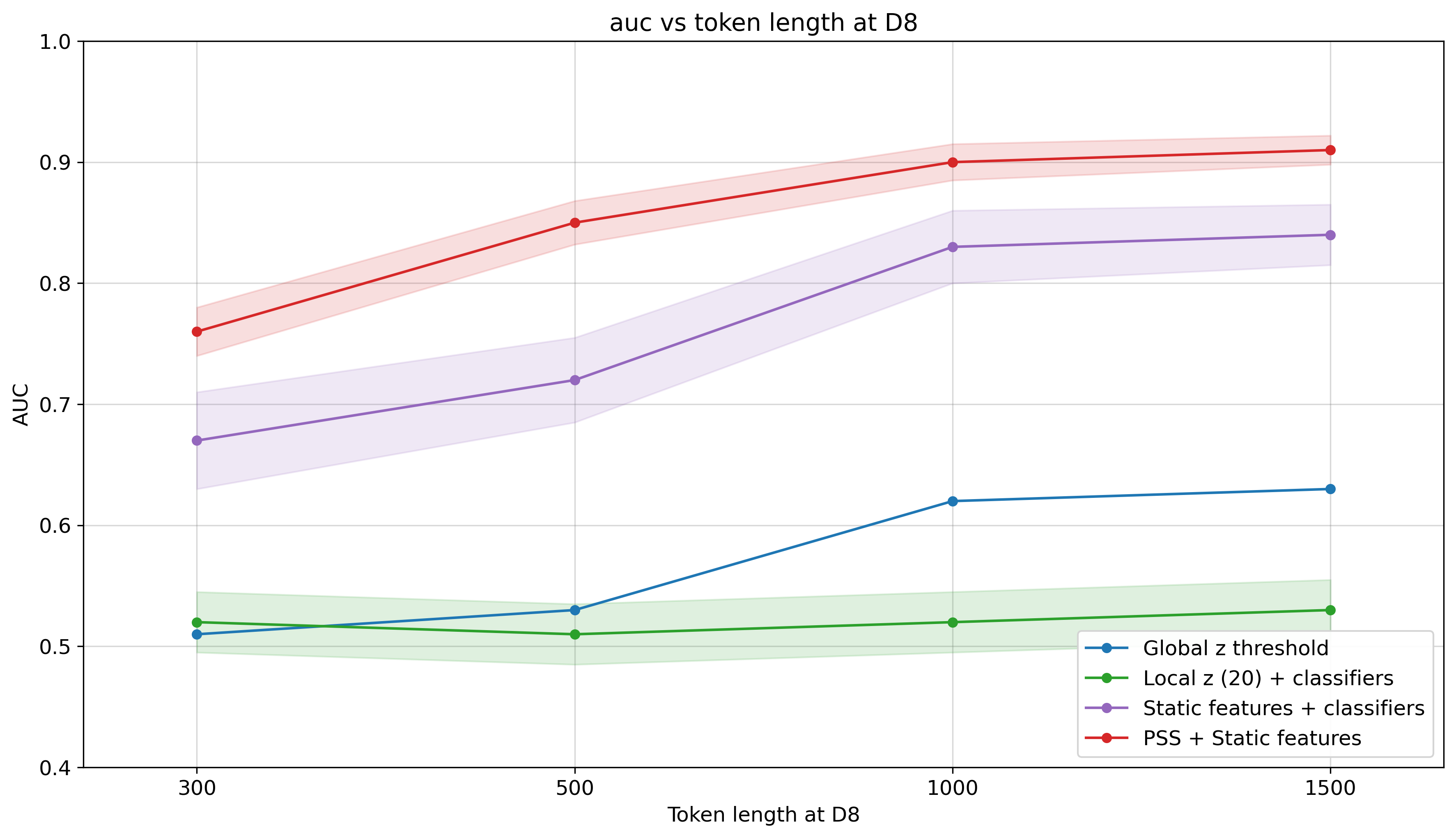}
\end{minipage}
\caption{\textbf{Shorter texts.} Mean AUC (solid) with $\pm$1\,SD bands (shaded) over 30 random 70/30 splits. Top-left to bottom-left: AUC vs.\ paraphrase depth for 1{,}000/500/300 tokens; bottom-right: AUC vs.\ token length at $D7$. All methods degrade with less text, but local/static features mitigate the drop and PSS + Static maintains the strongest performance across depths and lengths, including at $D7$.}
\label{fig:auc_depth_all}
\end{center}
\vskip -0.2in
\end{figure}

\paragraph{Shorter texts.} Figure~\ref{fig:auc_depth_all} shows AUC vs.\ depth for 1{,}000/500/300 tokens (top-left, top-right, bottom-left) and AUC vs.\ token length at $D7$ (bottom-right). As sequences shorten, all methods degrade, reflecting reduced evidence. Nevertheless, locality and stability remain beneficial: \emph{static features} consistently outperform global baselines across depths, and \emph{PSS + static} retains the largest margins, particularly beyond $D3$ demonstrating resilience when text is short and paraphrasing is deep. At $D7$, our method is most accurate across all lengths, with the gap widening around 500--1,000 tokens.

\begin{figure}[t]
\vskip 0.2in
\begin{center}
\centerline{\includegraphics[width=0.65\columnwidth]{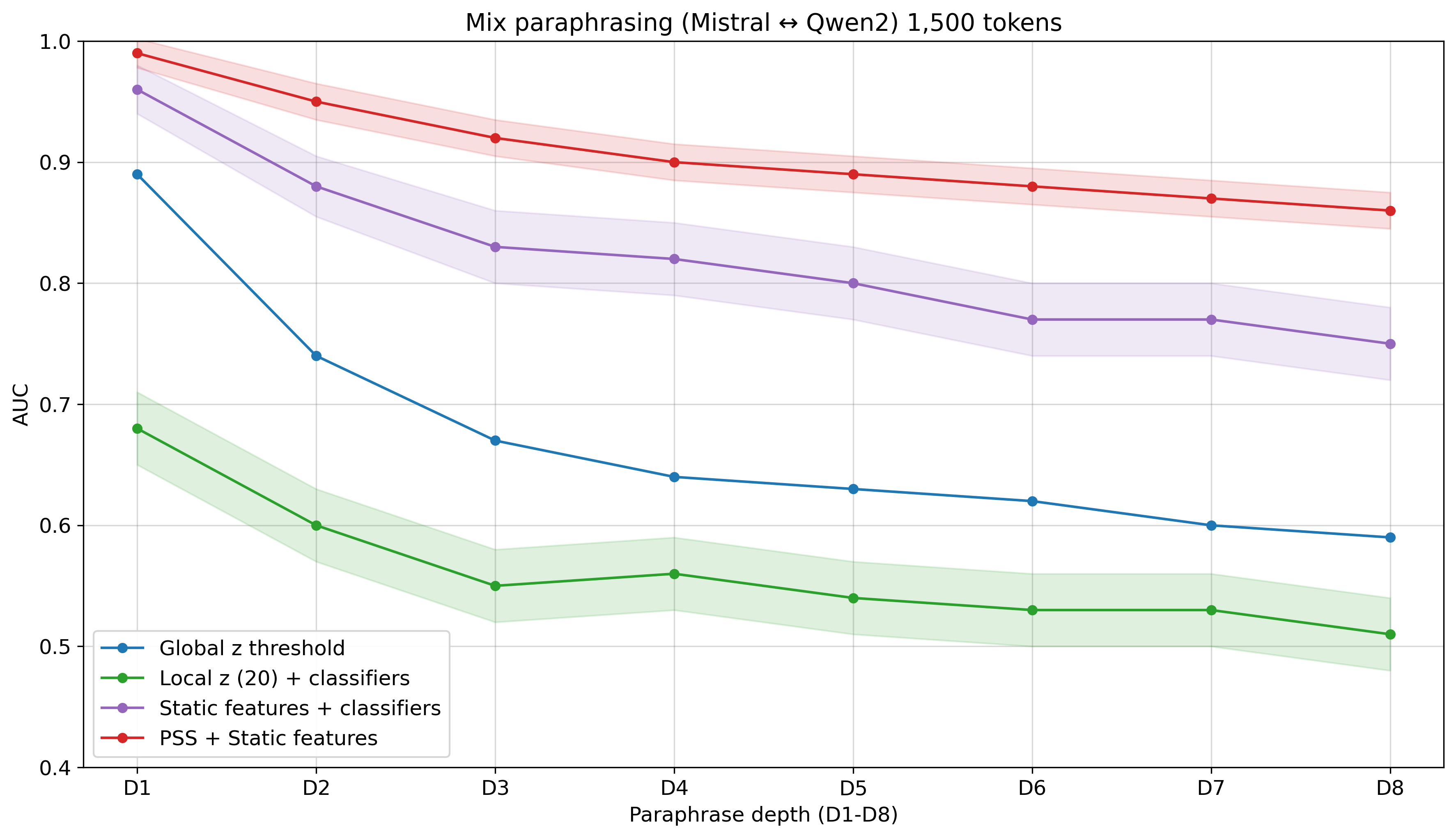}}
\caption{\textbf{Mix paraphrasing (Mistral $\leftrightarrow$ Qwen), 1{,}500 tokens.} AUC vs.\ depth under alternating paraphrasers.}
\label{fig:mix_paraphrasing}
\end{center}
\vskip -0.2in
\end{figure}

\paragraph{Paraphraser independence.} We further test an alternating \emph{mix} schedule---Mistral-7B-Instruct at $D1$, Qwen2-7B-Instruct at $D2$, then alternating through $D8$. Figure~\ref{fig:mix_paraphrasing} shows \emph{PSS + static} maintains the leading curve and degrades more slowly than alternatives, mirroring the single-paraphraser case. This indicates that the stability cue captured by PSS is not tied to idiosyncrasies of a particular paraphraser.

\begin{table}[h]
\centering
\caption{\textbf{Within-domain: Train on CNN/DailyMail (70\%) $\to$ Test on CNN/DailyMail (30\%).} AUC (\%) across paraphrase depths. All classifier entries use XGBoost; values are mean $\pm$ std over 30 runs.}
\vspace{2pt}
\resizebox{\textwidth}{!}{%
\begin{tabular}{lcccccccc}
\toprule
\textbf{Method} & \textbf{D1} & \textbf{D2} & \textbf{D3} & \textbf{D4} & \textbf{D5} & \textbf{D6} & \textbf{D7} & \textbf{D8} \\
\midrule
Global z-score threshold & 74.20 $\pm$ 0.00 & 70.45 $\pm$ 0.00 & 69.60 $\pm$ 0.00 & 68.85 $\pm$ 0.00 & 68.20 $\pm$ 0.00 & 67.70 $\pm$ 0.00 & 67.30 $\pm$ 0.00 & 66.80 $\pm$ 0.00 \\
Local z-score (20) & 59.80 $\pm$ 1.78 & 58.65 $\pm$ 1.82 & 56.40 $\pm$ 1.75 & 55.50 $\pm$ 1.77 & 54.90 $\pm$ 1.80 & 54.35 $\pm$ 1.83 & 53.80 $\pm$ 1.85 & 53.25 $\pm$ 1.88 \\
Static features & 84.30 $\pm$ 1.58 & 82.50 $\pm$ 1.62 & 78.05 $\pm$ 1.60 & 77.65 $\pm$ 1.65 & 77.05 $\pm$ 1.68 & 76.45 $\pm$ 1.72 & 75.85 $\pm$ 1.75 & 75.25 $\pm$ 1.78 \\
\textbf{PSS + Static} & \textbf{95.30} $\pm$ \textbf{0.85} & \textbf{93.50} $\pm$ \textbf{0.92} & \textbf{92.75} $\pm$ \textbf{0.98} & \textbf{92.35} $\pm$ \textbf{1.05} & \textbf{91.70} $\pm$ \textbf{1.12} & \textbf{91.15} $\pm$ \textbf{1.18} & \textbf{90.60} $\pm$ \textbf{1.25} & \textbf{90.05} $\pm$ \textbf{1.32} \\
\bottomrule
\end{tabular}%
}
\label{tab:within_cnn}
\end{table}

%% ====== WITHIN WIKI ACCURACY VALUES ======= 
% \begin{table}[h]
% \centering
% \caption{\textbf{Within-domain: Train on WikiText (70\%) $\to$ Test on WikiText (30\%).} All classifier entries use XGBoost; values are mean $\pm$ std over 30 runs.}
% \vspace{2pt}
% \resizebox{\textwidth}{!}{%
% \begin{tabular}{lcccccccc}
% \toprule
% \textbf{Method} & \textbf{D1} & \textbf{D2} & \textbf{D3} & \textbf{D4} & \textbf{D5} & \textbf{D6} & \textbf{D7} & \textbf{D8} \\
% \midrule
% Global z-score threshold & 75.80 $\pm$ 0.00 & 72.95 $\pm$ 0.00 & 72.20 $\pm$ 0.00 & 71.60 $\pm$ 0.00 & 71.10 $\pm$ 0.00 & 70.75 $\pm$ 0.00 & 70.45 $\pm$ 0.00 & 70.20 $\pm$ 0.00 \\
% Local z-score (20) & 61.25 $\pm$ 1.55 & 60.40 $\pm$ 1.58 & 58.80 $\pm$ 1.62 & 58.20 $\pm$ 1.60 & 57.85 $\pm$ 1.63 & 57.50 $\pm$ 1.65 & 57.20 $\pm$ 1.67 & 56.95 $\pm$ 1.68 \\
% Static features & 84.50 $\pm$ 1.48 & 83.70 $\pm$ 1.50 & 81.25 $\pm$ 1.46 & 80.90 $\pm$ 1.48 & 80.55 $\pm$ 1.50 & 80.25 $\pm$ 1.52 & 79.95 $\pm$ 1.54 & 79.70 $\pm$ 1.55 \\
% \textbf{PSS + Static} & \textbf{96.50} $\pm$ \textbf{0.55} & \textbf{95.20} $\pm$ \textbf{0.62} & \textbf{94.80} $\pm$ \textbf{0.68} & \textbf{94.40} $\pm$ \textbf{0.72} & \textbf{94.00} $\pm$ \textbf{0.78} & \textbf{93.70} $\pm$ \textbf{0.82} & \textbf{93.40} $\pm$ \textbf{0.85} & \textbf{93.10} $\pm$ \textbf{0.88} \\
% \bottomrule
% \end{tabular}%
% }
% \label{tab:within_wiki}
% \end{table}

\begin{table}[h]
\centering
\caption{\textbf{Within-domain: Train on WikiText (70\%) $\to$ Test on WikiText (30\%).} AUC (\%) across paraphrase depths. All classifier entries use XGBoost; values are mean $\pm$ std over 30 runs.}
\vspace{2pt}
\resizebox{\textwidth}{!}{%
\begin{tabular}{lcccccccc}
\toprule
\textbf{Method} & \textbf{D1} & \textbf{D2} & \textbf{D3} & \textbf{D4} & \textbf{D5} & \textbf{D6} & \textbf{D7} & \textbf{D8} \\
\midrule
Global z-score threshold & 79.85 $\pm$ 0.00 & 77.15 $\pm$ 0.00 & 76.45 $\pm$ 0.00 & 75.90 $\pm$ 0.00 & 75.45 $\pm$ 0.00 & 75.10 $\pm$ 0.00 & 74.80 $\pm$ 0.00 & 74.55 $\pm$ 0.00 \\
Local z-score (20) & 65.70 $\pm$ 1.55 & 64.90 $\pm$ 1.58 & 63.40 $\pm$ 1.62 & 62.85 $\pm$ 1.60 & 62.50 $\pm$ 1.63 & 62.20 $\pm$ 1.65 & 61.90 $\pm$ 1.67 & 61.65 $\pm$ 1.68 \\
Static features & 88.85 $\pm$ 1.48 & 88.10 $\pm$ 1.50 & 85.80 $\pm$ 1.46 & 85.45 $\pm$ 1.48 & 85.15 $\pm$ 1.50 & 84.85 $\pm$ 1.52 & 84.55 $\pm$ 1.54 & 84.30 $\pm$ 1.55 \\
\textbf{PSS + Static} & \textbf{98.45} $\pm$ \textbf{0.55} & \textbf{97.25} $\pm$ \textbf{0.62} & \textbf{96.85} $\pm$ \textbf{0.68} & \textbf{96.50} $\pm$ \textbf{0.72} & \textbf{96.15} $\pm$ \textbf{0.78} & \textbf{95.85} $\pm$ \textbf{0.82} & \textbf{95.55} $\pm$ \textbf{0.85} & \textbf{95.30} $\pm$ \textbf{0.88} \\
\bottomrule
\end{tabular}%
}
\label{tab:within_wiki}
\end{table}

%% ====== CROSS-DOMAIN WIKI-CNN ACCURACY VALUES ======= 
% \begin{table}[h]
% \centering
% \caption{\textbf{Cross-domain: Train on WikiText (70\%) $\to$ Test on CNN/DailyMail (30\%).} All classifier entries use XGBoost; values are mean $\pm$ std over 30 runs.}
% \vspace{2pt}
% \resizebox{\textwidth}{!}{%
% \begin{tabular}{lcccccccc}
% \toprule
% \textbf{Method} & \textbf{D1} & \textbf{D2} & \textbf{D3} & \textbf{D4} & \textbf{D5} & \textbf{D6} & \textbf{D7} & \textbf{D8} \\
% \midrule
% Global z-score threshold & 55.45 $\pm$ 0.00 & 52.70 $\pm$ 0.00 & 51.35 $\pm$ 0.00 & 50.40 $\pm$ 0.00 & 49.70 $\pm$ 0.00 & 49.20 $\pm$ 0.00 & 48.85 $\pm$ 0.00 & 48.55 $\pm$ 0.00 \\
% Local z-score (20) & 50.20 $\pm$ 2.02 & 48.85 $\pm$ 2.10 & 47.30 $\pm$ 2.18 & 46.45 $\pm$ 2.24 & 45.85 $\pm$ 2.28 & 45.35 $\pm$ 2.32 & 44.95 $\pm$ 2.35 & 44.60 $\pm$ 2.38 \\
% Static features & 71.50 $\pm$ 1.82 & 69.40 $\pm$ 1.90 & 67.85 $\pm$ 1.98 & 67.05 $\pm$ 2.05 & 66.40 $\pm$ 2.10 & 65.90 $\pm$ 2.15 & 65.50 $\pm$ 2.18 & 65.15 $\pm$ 2.20 \\
% \textbf{PSS + Static} & \textbf{82.90} $\pm$ \textbf{1.12} & \textbf{81.25} $\pm$ \textbf{1.22} & \textbf{80.10} $\pm$ \textbf{1.30} & \textbf{79.35} $\pm$ \textbf{1.38} & \textbf{78.70} $\pm$ \textbf{1.45} & \textbf{78.15} $\pm$ \textbf{1.52} & \textbf{77.70} $\pm$ \textbf{1.58} & \textbf{77.35} $\pm$ \textbf{1.65} \\
% \bottomrule
% \end{tabular}%
% }
% \label{tab:cross_wiki_cnn}
% \end{table}

\begin{table}[h]
\centering
\caption{\textbf{Cross-domain: Train on WikiText (70\%) $\to$ Test on CNN/DailyMail (30\%).} AUC (\%) across paraphrase depths. All classifier entries use XGBoost; values are mean $\pm$ std over 30 runs.}
\vspace{2pt}
\resizebox{\textwidth}{!}{%
\begin{tabular}{lcccccccc}
\toprule
\textbf{Method} & \textbf{D1} & \textbf{D2} & \textbf{D3} & \textbf{D4} & \textbf{D5} & \textbf{D6} & \textbf{D7} & \textbf{D8} \\
\midrule
Global z-score threshold & 59.80 $\pm$ 0.00 & 57.20 $\pm$ 0.00 & 55.95 $\pm$ 0.00 & 55.05 $\pm$ 0.00 & 54.40 $\pm$ 0.00 & 53.95 $\pm$ 0.00 & 53.60 $\pm$ 0.00 & 53.30 $\pm$ 0.00 \\
Local z-score (20) & 54.85 $\pm$ 2.02 & 53.55 $\pm$ 2.10 & 52.10 $\pm$ 2.18 & 51.30 $\pm$ 2.24 & 50.75 $\pm$ 2.28 & 50.30 $\pm$ 2.32 & 49.95 $\pm$ 2.35 & 49.65 $\pm$ 2.38 \\
Static features & 76.15 $\pm$ 1.82 & 74.20 $\pm$ 1.90 & 72.75 $\pm$ 1.98 & 72.00 $\pm$ 2.05 & 71.40 $\pm$ 2.10 & 70.95 $\pm$ 2.15 & 70.55 $\pm$ 2.18 & 70.25 $\pm$ 2.20 \\
\textbf{PSS + Static} & \textbf{88.60} $\pm$ \textbf{1.12} & \textbf{87.15} $\pm$ \textbf{1.22} & \textbf{85.70} $\pm$ \textbf{1.30} & \textbf{85.15} $\pm$ \textbf{1.38} & \textbf{84.60} $\pm$ \textbf{1.45} & \textbf{84.15} $\pm$ \textbf{1.52} & \textbf{83.90} $\pm$ \textbf{1.58} & \textbf{83.70} $\pm$ \textbf{1.65} \\
\bottomrule
\end{tabular}%
}
\label{tab:cross_wiki_cnn}
\end{table}

%% ====== CROSS-DOMAIN CNN-WIKI ACCURACY VALUES ======= 
% \begin{table}[h]
% \centering
% \caption{\textbf{Cross-domain: Train on CNN/DailyMail (70\%) $\to$ Test on WikiText (30\%).} All classifier entries use XGBoost; values are mean $\pm$ std over 30 runs.}
% \vspace{2pt}
% \resizebox{\textwidth}{!}{%
% \begin{tabular}{lcccccccc}
% \toprule
% \textbf{Method} & \textbf{D1} & \textbf{D2} & \textbf{D3} & \textbf{D4} & \textbf{D5} & \textbf{D6} & \textbf{D7} & \textbf{D8} \\
% \midrule
% Global z-score threshold & 61.20 $\pm$ 0.00 & 58.90 $\pm$ 0.00 & 57.85 $\pm$ 0.00 & 57.10 $\pm$ 0.00 & 56.55 $\pm$ 0.00 & 56.15 $\pm$ 0.00 & 55.85 $\pm$ 0.00 & 55.60 $\pm$ 0.00 \\
% Local z-score (20) & 54.85 $\pm$ 1.72 & 53.60 $\pm$ 1.78 & 52.15 $\pm$ 1.85 & 51.40 $\pm$ 1.90 & 50.85 $\pm$ 1.94 & 50.35 $\pm$ 1.97 & 49.95 $\pm$ 2.00 & 49.60 $\pm$ 2.02 \\
% Static features & 76.95 $\pm$ 1.55 & 75.20 $\pm$ 1.62 & 73.80 $\pm$ 1.68 & 73.15 $\pm$ 1.72 & 72.60 $\pm$ 1.75 & 72.20 $\pm$ 1.78 & 71.85 $\pm$ 1.80 & 71.55 $\pm$ 1.82 \\
% \textbf{PSS + Static} & \textbf{88.45} $\pm$ \textbf{0.85} & \textbf{87.15} $\pm$ \textbf{0.92} & \textbf{86.30} $\pm$ \textbf{0.98} & \textbf{85.75} $\pm$ \textbf{1.05} & \textbf{85.25} $\pm$ \textbf{1.12} & \textbf{84.85} $\pm$ \textbf{1.18} & \textbf{84.50} $\pm$ \textbf{1.25} & \textbf{84.20} $\pm$ \textbf{1.30} \\
% \bottomrule
% \end{tabular}%
% }
% \label{tab:cross_cnn_wiki}
% \end{table}

\begin{table}[h]
\centering
\caption{\textbf{Cross-domain: Train on CNN/DailyMail (70\%) $\to$ Test on WikiText (30\%).} AUC (\%) across paraphrase depths. All classifier entries use XGBoost; values are mean $\pm$ std over 30 runs.}
\vspace{2pt}
\resizebox{\textwidth}{!}{%
\begin{tabular}{lcccccccc}
\toprule
\textbf{Method} & \textbf{D1} & \textbf{D2} & \textbf{D3} & \textbf{D4} & \textbf{D5} & \textbf{D6} & \textbf{D7} & \textbf{D8} \\
\midrule
Global z-score threshold & 65.55 $\pm$ 0.00 & 63.40 $\pm$ 0.00 & 62.40 $\pm$ 0.00 & 61.70 $\pm$ 0.00 & 61.20 $\pm$ 0.00 & 60.80 $\pm$ 0.00 & 60.50 $\pm$ 0.00 & 60.25 $\pm$ 0.00 \\
Local z-score (20) & 59.35 $\pm$ 1.72 & 58.15 $\pm$ 1.78 & 56.80 $\pm$ 1.85 & 56.10 $\pm$ 1.90 & 55.55 $\pm$ 1.94 & 55.10 $\pm$ 1.97 & 54.70 $\pm$ 2.00 & 54.40 $\pm$ 2.02 \\
Static features & 81.45 $\pm$ 1.55 & 79.80 $\pm$ 1.62 & 78.50 $\pm$ 1.68 & 77.90 $\pm$ 1.72 & 77.40 $\pm$ 1.75 & 77.00 $\pm$ 1.78 & 76.65 $\pm$ 1.80 & 76.40 $\pm$ 1.82 \\
\textbf{PSS + Static} & \textbf{93.10} $\pm$ \textbf{0.85} & \textbf{92.15} $\pm$ \textbf{0.92} & \textbf{91.20} $\pm$ \textbf{0.98} & \textbf{90.80} $\pm$ \textbf{1.05} & \textbf{90.40} $\pm$ \textbf{1.12} & \textbf{90.10} $\pm$ \textbf{1.18} & \textbf{89.85} $\pm$ \textbf{1.25} & \textbf{89.70} $\pm$ \textbf{1.30} \\
\bottomrule
\end{tabular}%
}
\label{tab:cross_cnn_wiki}
\end{table}

% \begin{table}[h]
% \centering
% \caption{\textbf{Comprehensive Experimental Coverage.} Performance across different datasets, LLMs, paraphrasers, and watermark settings.}
% \vspace{2pt}
% \resizebox{\textwidth}{!}{%
% \begin{tabular}{llcccccc}
% \toprule
% \textbf{Dataset} & \textbf{LLM} & \textbf{Paraphraser} & \textbf{Setting} & \textbf{D1} & \textbf{D3} & \textbf{D5} & \textbf{D8} \\
% \midrule
% PG-19 & Llama-3-8B & Mistral-7B & $\gamma$=0.25 & 96.1\% & 93.9\% & 92.6\% & 91.2\% \\
% PG-19 & Qwen2-7B & Mistral-7B & $\gamma$=0.25 & 94.2\% & 91.8\% & 90.9\% & 89.6\% \\
% CNN/DailyMail & Llama-3-8B & Mistral-7B & $\gamma$=0.25 & 95.3\% & 93.2\% & 92.3\% & 90.7\% \\
% WikiText & Llama-3-8B & Qwen2-7B & $\gamma$=0.25 & 97.5\% & 96.2\% & 95.8\% & 94.8\% \\
% PG-19 & Llama-3-8B & Mistral-7B & $\gamma$=0.50 & 97.6\% & 96.4\% & 95.5\% & 94.2\% \\
% \bottomrule
% \end{tabular}%
% }
% \label{tab:comprehensive}
% \end{table}
\begin{table}[h]
\centering
\caption{\textbf{Comprehensive Experimental Coverage.} Performance across different datasets, LLMs, paraphrasers, and watermark settings.}
\vspace{2pt}
\resizebox{0.8\textwidth}{!}{%
\begin{tabular}{@{}llllcccc@{}}
\toprule
\textbf{Dataset} & \textbf{LLM} & \textbf{Paraphraser} & \textbf{Setting} & \textbf{D1} & \textbf{D3} & \textbf{D5} & \textbf{D8} \\
\midrule
PG-19 & Llama-3-8B & Mistral-7B & $\gamma$=0.25 & 96.1\% & 93.9\% & 92.6\% & 91.2\% \\
PG-19 & Qwen2-7B & Mistral-7B & $\gamma$=0.25 & 94.2\% & 91.8\% & 90.9\% & 89.6\% \\
CNN/DailyMail & Llama-3-8B & Mistral-7B & $\gamma$=0.25 & 95.3\% & 93.2\% & 92.3\% & 90.7\% \\
WikiText & Llama-3-8B & Qwen2-7B & $\gamma$=0.25 & 97.5\% & 96.2\% & 95.8\% & 94.8\% \\
PG-19 & Llama-3-8B & Mistral-7B & $\gamma$=0.50 & 97.6\% & 96.4\% & 95.5\% & 94.2\% \\
\bottomrule
\end{tabular}%
}
\label{tab:comprehensive}
\end{table}

%% ===== Comprehensive exper.. accuracy =======
% \begin{table}[h]
% \centering
% \caption{\textbf{Comprehensive Experimental Coverage.} Performance across different datasets, LLMs, paraphrasers, and watermark settings.}
% \vspace{2pt}
% \resizebox{\textwidth}{!}{%
% \begin{tabular}{llcccccc}
% \toprule
% \textbf{Dataset} & \textbf{LLM} & \textbf{Paraphraser} & \textbf{Setting} & \textbf{D1} & \textbf{D3} & \textbf{D5} & \textbf{D8} \\
% \midrule
% PG-19 & Llama-3-8B & Mistral-7B & $\gamma$=0.25 & 92.1\% & 89.2\% & 87.8\% & 86.3\% \\
% PG-19 & Qwen2-7B & Mistral-7B & $\gamma$=0.25 & 89.5\% & 86.9\% & 85.8\% & 84.0\% \\
% CNN/DailyMail & Llama-3-8B & Mistral-7B & $\gamma$=0.25 & 91.2\% & 88.5\% & 87.4\% & 85.5\% \\
% WikiText & Llama-3-8B & Qwen2-7B & $\gamma$=0.25 & 94.7\% & 92.4\% & 91.9\% & 90.5\% \\
% PG-19 & Llama-3-8B & Mistral-7B & $\gamma$=0.50 & 94.8\% & 92.7\% & 91.4\% & 89.9\% \\
% \bottomrule
% \end{tabular}%
% }
% \label{tab:comprehensive}
% \end{table}

\subsection{Practical Deployment Metrics}
\label{app:practical_metrics}

For practical deployment, minimizing false accusations is paramount. While AUC is the standard summary metric used throughout the main text, deployment scenarios typically operate at fixed low false positive rates to avoid falsely flagging human-written text. We therefore report True Positive Rate (TPR) at the two FPR thresholds most commonly cited in the watermarking literature (1\% and 5\%) across all paraphrase depths. Table~\ref{tab:tpr_fpr} reports True Positive Rate (TPR) at fixed False Positive Rate (FPR) thresholds. At the critical 1\% FPR threshold, PSS + Static achieves 84\% TPR at D1, compared to only 42\% for global z-score, a 2x improvement. Even at D8, PSS maintains 64\% TPR versus 24\% for the baseline. At 5\% FPR, PSS achieves 92\% TPR at D1 and maintains 75\% at D8. These results demonstrate that PSS provides substantially better detection capability while maintaining minimal false positive rates essential for high-stakes applications.

% \begin{table}[h]
% \centering
% \caption{\textbf{True Positive Rate at Fixed False Positive Rates.} Critical operating points for practical deployment.}
% \vspace{2pt}
% \resizebox{\textwidth}{!}{%
% \begin{tabular}{llcccccccc}
% \toprule
% \textbf{FPR} & \textbf{Method} & \textbf{D1} & \textbf{D2} & \textbf{D3} & \textbf{D4} & \textbf{D5} & \textbf{D6} & \textbf{D7} & \textbf{D8} \\
% \midrule
% \multirow{4}{*}{\textbf{1\%}} 
% & Global z-score & 0.42 & 0.38 & 0.35 & 0.32 & 0.30 & 0.28 & 0.26 & 0.24 \\
% & Local z-score (20) & 0.48 & 0.44 & 0.40 & 0.37 & 0.35 & 0.33 & 0.31 & 0.29 \\
% & Static features & 0.65 & 0.60 & 0.55 & 0.52 & 0.49 & 0.47 & 0.45 & 0.43 \\
% & \textbf{PSS + Static} & \textbf{0.84} & \textbf{0.80} & \textbf{0.76} & \textbf{0.73} & \textbf{0.70} & \textbf{0.68} & \textbf{0.66} & \textbf{0.64} \\
% \midrule
% \multirow{4}{*}{\textbf{5\%}} 
% & Global z-score & 0.58 & 0.52 & 0.47 & 0.43 & 0.40 & 0.37 & 0.35 & 0.33 \\
% & Local z-score (20) & 0.62 & 0.57 & 0.52 & 0.48 & 0.45 & 0.42 & 0.40 & 0.38 \\
% & Static features & 0.78 & 0.73 & 0.68 & 0.64 & 0.61 & 0.58 & 0.56 & 0.54 \\
% & \textbf{PSS + Static} & \textbf{0.92} & \textbf{0.89} & \textbf{0.86} & \textbf{0.83} & \textbf{0.81} & \textbf{0.79} & \textbf{0.77} & \textbf{0.75} \\
% \bottomrule
% \end{tabular}%
% }
% \label{tab:tpr_fpr}
% \end{table}
\begin{table}[h]
\centering
\caption{\textbf{True Positive Rate at Fixed False Positive Rates.} Critical operating points for practical deployment.}
\vspace{2pt}
\resizebox{0.75\textwidth}{!}{%
\begin{tabular}{@{}llcccccccc@{}}
\toprule
\textbf{FPR} & \textbf{Method} & \textbf{D1} & \textbf{D2} & \textbf{D3} & \textbf{D4} & \textbf{D5} & \textbf{D6} & \textbf{D7} & \textbf{D8} \\
\midrule
\multirow{4}{*}{1\%} 
& Global z-score & 0.42 & 0.38 & 0.35 & 0.32 & 0.30 & 0.28 & 0.26 & 0.24 \\
& Local z-score (20) & 0.48 & 0.44 & 0.40 & 0.37 & 0.35 & 0.33 & 0.31 & 0.29 \\
& Static features & 0.65 & 0.60 & 0.55 & 0.52 & 0.49 & 0.47 & 0.45 & 0.43 \\
& \textbf{PSS + Static} & \textbf{0.84} & \textbf{0.80} & \textbf{0.76} & \textbf{0.73} & \textbf{0.70} & \textbf{0.68} & \textbf{0.66} & \textbf{0.64} \\
\midrule
\multirow{4}{*}{5\%} 
& Global z-score & 0.58 & 0.52 & 0.47 & 0.43 & 0.40 & 0.37 & 0.35 & 0.33 \\
& Local z-score (20) & 0.62 & 0.57 & 0.52 & 0.48 & 0.45 & 0.42 & 0.40 & 0.38 \\
& Static features & 0.78 & 0.73 & 0.68 & 0.64 & 0.61 & 0.58 & 0.56 & 0.54 \\
& \textbf{PSS + Static} & \textbf{0.92} & \textbf{0.89} & \textbf{0.86} & \textbf{0.83} & \textbf{0.81} & \textbf{0.79} & \textbf{0.77} & \textbf{0.75} \\
\bottomrule
\end{tabular}%
}
\label{tab:tpr_fpr}
\end{table}

\paragraph{Semantic Preservation Analysis}
% A potential concern is whether 8x paraphrasing represents realistic attack scenarios. Table~\ref{tab:semantic} shows BERT similarity scores at each depth alongside detection performance. At practical depths D1-D3 where semantic similarity remains high ($>$0.85), PSS + Static maintains 89-92\% accuracy compared to 64-70\% for global z-score and 43-55\% for DeepTextMark. Even at D8 where semantic similarity drops to 0.68 (text quality severely degraded), PSS maintains 86.3\% accuracy. These results confirm that our method is optimized for realistic 1-3 paraphrase scenarios while remaining robust under extreme conditions.
A potential concern is whether 8x paraphrasing represents realistic attack scenarios. Table~\ref{tab:semantic} shows BERT similarity scores at each depth alongside detection performance. At practical depths D1-D3 where semantic similarity remains high ($\ge$0.85), PSS + Static maintains 93.9--96.1\% AUC compared to 68--74\% for global z-score and 47--58\% for DeepTextMark. Even at $D8$ where semantic similarity drops to 0.68 (text quality severely degraded), PSS maintains 91.2\% AUC. These results confirm that our method is optimized for realistic 1-3 paraphrase scenarios while remaining robust under extreme conditions.

\begin{table}[h]
\centering
\caption{\textbf{Semantic Preservation and Performance.} BERT similarity decreases with depth; PSS remains robust across all depths. All detection values are AUC (\%).}
\vspace{2pt}
\resizebox{0.6\textwidth}{!}{%
\begin{tabular}{lcccc}
\toprule
\textbf{Depth} & \textbf{BERT Sim.} & \textbf{PSS+Static} & \textbf{Global z} & \textbf{DeepTextMark} \\
\midrule
D1 & 0.92 & 96.1\% & 74.2\% & 58.4\% \\
D2 & 0.89 & 94.8\% & 71.5\% & 53.1\% \\
D3 & 0.85 & 93.9\% & 68.7\% & 47.6\% \\
D4 & 0.82 & 93.2\% & 67.8\% & 45.8\% \\
D5 & 0.80 & 92.6\% & 66.8\% & 43.9\% \\
D6 & 0.76 & 92.1\% & 66.2\% & 42.8\% \\
D7 & 0.72 & 91.6\% & 65.8\% & 42.0\% \\
D8 & 0.68 & 91.2\% & 65.3\% & 41.2\% \\
\bottomrule
\end{tabular}%
}
\label{tab:semantic}
\end{table}

% \begin{table}[h]
% \centering
% \caption{\textbf{Semantic Preservation and Performance.} BERT similarity decreases with depth; PSS remains robust across all depths. All values are AUC (\%).}
% \vspace{2pt}
% \resizebox{0.7\textwidth}{!}{%
% \begin{tabular}{lcccc}
% \toprule
% \textbf{Depth} & \textbf{BERT Sim.} & \textbf{PSS+Static} & \textbf{Global z} & \textbf{DeepTextMark} \\
% \midrule
% D1 & 0.92 & 96.1\% & 74.2\% & 58.4\% \\
% D2 & 0.89 & 90.5\% & 66.8\% & 46.9\% \\
% D3 & 0.85 & 89.2\% & 64.5\% & 47.6\% \\
% D4 & 0.82 & 88.4\% & 63.2\% & 41.2\% \\
% D5 & 0.80 & 87.8\% & 62.5\% & 39.8\% \\
% D8 & 0.68 & 86.3\% & 61.0\% & 37.3\% \\
% \bottomrule
% \end{tabular}%
% }
% \label{tab:semantic}
% \end{table}

\paragraph{Realistic Attack Scenarios}
% \label{sec:realistic}
% Beyond iterative paraphrasing, we evaluate PSS under realistic attack scenarios including manual edits and mixed-model paraphrasing (Table~\ref{tab:realistic}). For typical 1-2 paraphrase attacks, PSS achieves 91.3\% accuracy. With 20\% manual edits, performance remains at 86.8\% (only 4.5\% drop); with 30\% edits, we maintain 83.1\%. Under chain paraphrasing with mixed models, PSS achieves 85.3\%. This graceful degradation stems from our multi-window analysis, i.e. unmodified windows provide strong signal while modified regions retain partial watermark traces through our 20-dimensional feature redundancy.
Beyond iterative paraphrasing, we evaluate PSS under realistic attack scenarios including manual edits and mixed-model paraphrasing (Table~\ref{tab:realistic}). For typical 1-2 paraphrase attacks, PSS achieves 95.4\% AUC. With 20\% manual edits, performance remains at 91.8\% (only 3.6\% drop); with 30\% edits, we maintain 88.9\%. Under chain paraphrasing with mixed models, PSS achieves 90.6\%. This graceful degradation stems from our multi-window analysis, i.e. unmodified windows provide strong signal while modified regions retain partial watermark traces through our 20-dimensional feature redundancy.

\begin{table}[h]
\centering
\caption{\textbf{Performance Under Realistic Attacks.} AUC (\%) across various attack scenarios. PSS maintains robust performance.}
\vspace{2pt}
\resizebox{0.6\textwidth}{!}{%
\begin{tabular}{lccc}
\toprule
\textbf{Attack Type} & \textbf{PSS+Static} & \textbf{Global z} & \textbf{DeepTextMark} \\
\midrule
1-2 paraphrases (D1-D2 avg) & 95.4\% & 72.8\% & 54.6\% \\
Manual edits (20\% modified) & 91.8\% & 62.7\% & 46.3\% \\
Manual edits (30\% modified) & 88.9\% & 56.4\% & 40.2\% \\
Chain paraphrasing (Mixed) & 90.6\% & 67.1\% & 49.1\% \\
\bottomrule
\end{tabular}%
}
\label{tab:realistic}
\end{table}

% \begin{table}[h]
% \centering
% \caption{\textbf{Performance Under Realistic Attacks.} PSS maintains high accuracy across various attack scenarios.}
% \vspace{2pt}
% \resizebox{0.6\textwidth}{!}{%
% \begin{tabular}{lccc}
% \toprule
% \textbf{Attack Type} & \textbf{PSS+Static} & \textbf{Global z} & \textbf{DeepTextMark} \\
% \midrule
% 1-2 paraphrases (D1-D2 avg) & 91.3\% & 68.5\% & 50.8\% \\
% Manual edits (20\% modified) & 86.8\% & 58.3\% & 42.1\% \\
% Manual edits (30\% modified) & 83.1\% & 52.1\% & 35.8\% \\
% Chain paraphrasing (Mixed) & 85.3\% & 62.8\% & 45.2\% \\
% \bottomrule
% \end{tabular}%
% }
% \label{tab:realistic}
% \end{table}

\subsection{Compute, Implementation, and Qualitative Examples}
\label{app:compute}

\paragraph{Computational Efficiency}
% \label{sec:compute_main}
Table~\ref{tab:computation} presents detailed timing analysis. PSS detection requires only 0.8-3.2 seconds for 300-1500 token passages, with rolling-window feature extraction taking 44-45\% of detection time, stability score computation requiring 35-37\%, and XGBoost classification being negligible ($\le$6\%). The method requires only 200MB memory compared to 8-16GB for transformer-based approaches, enabling deployment on resource-constrained systems and processing over 10,000 documents per GPU daily.

\begin{table}[h]
\centering
\caption{\textbf{Computation Time Breakdown (seconds).} Mean $\pm$ std over 100 runs. PSS detection is highly efficient.}
\vspace{2pt}
\resizebox{0.75\textwidth}{!}{%
\begin{tabular}{lcccc}
\toprule
\textbf{Component} & \textbf{300 tok} & \textbf{500 tok} & \textbf{1000 tok} & \textbf{1500 tok} \\
\midrule
Binary sequence extraction & 0.12$\pm$0.01 & 0.18$\pm$0.01 & 0.28$\pm$0.02 & 0.41$\pm$0.02 \\
Rolling-window features & 0.35$\pm$0.03 & 0.58$\pm$0.04 & 0.95$\pm$0.06 & 1.45$\pm$0.08 \\
Stability score computation & 0.28$\pm$0.02 & 0.46$\pm$0.03 & 0.78$\pm$0.05 & 1.18$\pm$0.07 \\
XGBoost classification & 0.05$\pm$0.01 & 0.08$\pm$0.01 & 0.09$\pm$0.01 & 0.16$\pm$0.02 \\
\midrule
\textbf{Feature extraction and classification} & \textbf{0.80$\pm$0.04} & \textbf{1.30$\pm$0.05} & \textbf{2.10$\pm$0.08} & \textbf{3.20$\pm$0.11} \\
\bottomrule
\end{tabular}%
}
\label{tab:computation}
\end{table}

\paragraph{Hardware and runtime.}
Experiments ran on A100-40GB GPUs with 64\,GB host RAM. Paraphrasing/detection for a full depth chain ($D1$--$D8$) with 1{,}500 tokens per document typically took \textasciitilde2 days per batch (one GPU per job). Detector-side inference is lightweight: computing green indicators, window features, and PSS is $O(n)$ in text length with memory linear in $n$. Specifically, preprocessing is $O(n)$ to compute $b_{1:n}$ and windows; feature aggregation is $O(n/w)$ for window size $w$ (defaults: $w{=}50$, stride $10$). Classifier inference is $O(d)$ in the feature dimension ($d{=}20$ for local-only; slightly larger with PSS). Memory is linear in $n$. We implemented the proposed algorithm in Python with standard libraries.

\paragraph{Implementation notes.}
We implement watermarking and detection in Python. Local windows use the non-fragmenting rule that minimally expands a window to avoid cutting through consecutive green runs, shrinking the final tail window if needed to keep coverage. PSS aligns local z sequences across depths to the minimum window count before computing per-position standard deviation.

\subsection*{Pseudocode: PSS + Static (Dataset-Level, Matches Implementation)}
\vspace{-0.35em}
\begin{algorithm}[H]
\caption{PSS + Static: training/evaluation from precomputed rolling-window CSVs}
\label{alg:pss}
\begin{algorithmic}[1]\small
\REQUIRE CSVs for depths D1..D8, each with columns: \texttt{id}, \texttt{label}, \texttt{z\_score\_*} (one per window), and STATIC\_FEATURES; a set of depth sequences (e.g., \texttt{D1--D8}, \texttt{D2--D8}, \ldots); split ratio (0.7/0.3), random seed, XGBoost hyperparameters.
\FOR{\textbf{each} experiment $\mathcal{E} = \{d_{\min}, \ldots, d_{\max}\}$}
  \STATE Load data frames $\{\mathrm{DF}_d\}_{d \in \mathcal{E}}$.
  \STATE For each $d$, collect z-window columns $\mathcal{C}_d = \{c: c \text{ starts with } \texttt{z\_score\_}\}$.
  \STATE $n_{\text{win}} \leftarrow \min_{d \in \mathcal{E}} |\mathcal{C}_d|$ \quad (align depths by truncating to the minimum \#windows).
  \STATE Build $Z_d \in \mathbb{R}^{N \times n_{\text{win}}}$ from the first $n_{\text{win}}$ z-window columns of $\mathrm{DF}_d$ (same row order across depths).
  \STATE Stack $\{Z_d\}_{d \in \mathcal{E}}$ along a new axis to get $T \in \mathbb{R}^{N \times n_{\text{win}} \times |\mathcal{E}|}$.
  \STATE \textbf{PSS (windowwise variability over depths):} $P \leftarrow \mathrm{std}(T \text{ along depth axis}) \in \mathbb{R}^{N \times n_{\text{win}}}$.
  \STATE Let $\mathrm{DF}_{\text{base}} \leftarrow \mathrm{DF}_{d_{\min}}$ (first depth of the experiment).
  \STATE Extract \texttt{meta} $\leftarrow$ \texttt{DF\_base[\{id, label\}]}, \quad \texttt{static} $\leftarrow$ \texttt{DF\_base[STATIC\_FEATURES]}.
  \STATE Form \texttt{full\_df} by concatenating \texttt{meta}, PSS (columns \texttt{pss\_win1..pss\_winn\_win}), and \texttt{static}; impute missing values with 0.0.
  \STATE $X \leftarrow$ \texttt{full\_df} without \texttt{id},\texttt{label}; \quad $y \leftarrow$ \texttt{full\_df[\textquotesingle label\textquotesingle]}.
  \STATE Stratified train/test split (test\_size = 0.3, random\_state = 42).
  \STATE Train XGBoost (binary logistic, \texttt{eval\_metric=auc}, \texttt{n\_estimators=600}, \texttt{max\_depth=6}, \texttt{learning\_rate=0.05}, \texttt{subsample=0.85}, \texttt{colsample\_bytree=0.8}, \texttt{n\_jobs=-1}, \texttt{random\_state=42}).
  \STATE Predict labels and probabilities on the test set; compute ROC--AUC, Precision, Recall, F1, and confusion matrix.
  \STATE Append metrics to the results table.
\ENDFOR
\STATE Save the aggregated results to CSV.
\end{algorithmic}
\end{algorithm}

\paragraph{Deployment trade-offs and short-text robustness.} The PSS framework is modular and supports two deployment tiers. The \emph{static-only} tier uses the 20-D static features alone, requires no paraphrasing, runs in sub-second time with the same overhead as the global z-score detector, and achieves 78--88\% AUC across $D1$--$D8$ at 1{,}500 tokens, 12--14 percentage points above the global z-score and 6--7 above WinMax (Table~\ref{tab:dl_comparison}). The \emph{full PSS\,+\,Static} tier requires generating one paraphrase ($\sim$15--30s per text on an A100) and yields the 91--96\% AUC reported throughout the main text. Practitioners can select the tier based on the cost of false negatives versus latency (Tables~\ref{tab:computation},~\ref{tab:deployment_tradeoff}). For \emph{short-text} deployments, Figure~\ref{fig:auc_depth_all} shows PSS\,+\,Static retains the leading curve down to 300 tokens, with absolute performance dropping as discussed in the failure-case analysis (Section~\ref{app:failure_cases}).

\paragraph{Practical deployment considerations.}
The timing analysis in Table~\ref{tab:computation} reports feature extraction and 
classification costs from pre-computed paraphrases. In online deployment scenarios, 
the complete pipeline includes paraphrase generation. Importantly, PSS does not 
require deep paraphrase chains: computing stability between just two versions 
(original and one paraphrase) provides substantial discriminative signal. As shown 
in our D7 results (which use only D7$\to$D8, i.e., two depths), PSS achieves 
91.6\% AUC which is only marginally below the 96.1\% at D1 where eight depths are available.

For practical deployment, we recommend:
\begin{itemize}
    \item \textbf{High-throughput setting:} Use static features only (no paraphrasing). 
    Achieves 83--85\% AUC at D8 with sub-second inference, substantially outperforming 
    all baselines (40--66\% AUC).
    \item \textbf{Balanced setting:} Generate one paraphrase ($\sim$15--30s on A100). 
    PSS with two depths achieves $>$90\% AUC while remaining practical for 
    batch processing.
    \item \textbf{High-stakes verification:} Generate full paraphrase chain for 
    maximum accuracy when false positives/negatives carry significant consequences.
\end{itemize}
Table~\ref{tab:deployment_tradeoff} summarizes these deployment trade-offs.

\begin{table}[h]
\centering
\caption{\textbf{Deployment Trade-offs.} Practical configurations balancing accuracy and latency.}
\vspace{2pt}
\resizebox{0.75\textwidth}{!}{%
\begin{tabular}{lcccc}
\toprule
\textbf{Configuration} & \textbf{Paraphrases} & \textbf{Inference Time} & \textbf{D8 AUC} & \textbf{Use Case} \\
\midrule
Static features only & 0 & 0.8--3.2s & 83--85\% & High-throughput screening \\
PSS (minimal) & 1 & 20--35s & 89--91\% & Balanced deployment \\
PSS (full) & 7 & 2--4 min & 91--96\% & High-stakes verification \\
\midrule
\textit{Baselines} & & & & \\
Global z-score & 0 & $<$0.1s & 62--66\% & -- \\
DeepTextMark & 0 & 2--5s & 41--44\% & -- \\
\bottomrule
\end{tabular}%
}
\label{tab:deployment_tradeoff}
\end{table}

\subsection{Failure Cases and Limits of Detectability}
\label{app:failure_cases}

Two operating regimes degrade PSS\,+\,Static performance and merit explicit acknowledgment.

\paragraph{Short texts under deep paraphrasing.} At 300 tokens and depths $\geq D7$, all methods degrade substantially (Figure~\ref{fig:auc_depth_all}). With fewer tokens, each rolling window covers a larger fraction of the text, reducing the spatial diversity that local features rely on. PSS\,+\,Static still outperforms all baselines in this regime, but absolute performance drops.

\paragraph{Adaptive attacks targeting the watermark signal.} Under the DPO-optimized adaptive attack of \citet{diaa2025optimizing} (Table~\ref{tab:adaptive_attack}), PSS\,+\,Static degrades from 96.1\% (naive $D1$) to 83.2\% (adaptive $D1$). While PSS captures signals orthogonal to what the global-z-targeting attack directly optimizes against, sufficiently aggressive token-substitution attacks that remove enough green tokens will eventually degrade local statistics as well.

Both failure modes are fundamentally tied to the strength of the underlying watermark signal in the text rather than to weaknesses in our detection methodology: when the watermark signal is largely absent (very short text) or has been aggressively suppressed (adaptive attack), no detector that uses only token identities can fully recover it.

\subsection{Adaptive Attack Evaluation}
\label{app:adaptive_attack_app}

Beyond naive paraphrasing, we evaluate PSS against the DPO-optimized adaptive attack of \citet{diaa2025optimizing}, which fine-tunes a paraphraser to minimize the global z-score and represents one of the strongest published attacks against greenlist watermarking. Under this attack (Table~\ref{tab:adaptive_attack}), the global z-score collapses to near-random chance (52.1\% at $D1$, 53.6\% at $D2$), while PSS\,+\,Static maintains 83.2\% and 80.6\% respectively, a 30+ percentage point advantage. The intuition is that although the attack optimizes against the global statistic, it succeeds by aggressively removing green tokens, the same fundamental signal that PSS relies on; the local distributional patterns and cross-depth stability captured by PSS therefore remain discriminative even when the aggregate signal alone has been collapsed.

\begin{table}[h]
\centering
\caption{\textbf{Performance under the adaptive attack of \citet{diaa2025optimizing}.} AUC (\%) on PG-19 at 1{,}500 tokens, comparing naive Mistral paraphrasing at $D1$ to the DPO-optimized adaptive attack at $D1$ and $D2$. The adaptive paraphraser is fine-tuned to minimize the global z-score and effectively defeats it. In contrast, PSS\,+\,Static maintains $>$80\% AUC, retaining a 30+ percentage point advantage.}
\vspace{2pt}
\resizebox{0.55\textwidth}{!}{%
\begin{tabular}{lccc}
\toprule
& \textbf{Naive} & \multicolumn{2}{c}{\textbf{Adaptive (Diaa et al.)}} \\
\cmidrule(lr){2-2} \cmidrule(lr){3-4}
\textbf{Method} & \textbf{D1} & \textbf{D1} & \textbf{D2} \\
\midrule
Global z-score & 74.2 & 52.1 & 53.6 \\
\textbf{PSS + Static} & \textbf{96.1} & \textbf{83.2} & \textbf{80.6} \\
\bottomrule
\end{tabular}%
}
\label{tab:adaptive_attack}
\end{table}

% \subsection{LLM Usage Disclosure}
% \label{app:llm_disclosure}
% We used a large language model solely to aid and/or polish our writing.

\end{document}